\documentclass[lettersize,journal]{IEEEtran}
\usepackage{amsmath,amsfonts,amssymb}
\usepackage{algorithmic}
\usepackage{array}
\usepackage{caption, subcaption}
\usepackage{textcomp}
\usepackage{stfloats}
\usepackage{url}
\usepackage{verbatim}
\usepackage{lettrine}
\usepackage{graphicx}
\usepackage{pifont}
\usepackage{caption}
\usepackage{graphicx}
\usepackage{xcolor}
\usepackage{float} 
\usepackage{subcaption}
\usepackage{colortbl}
\usepackage{multirow}
\usepackage{booktabs}
\usepackage{soul}
\usepackage{cite}
\usepackage{bm}
\def\BibTeX{{\mathrm B\kern-.05em{\sc i\kern-.025em b}\kern-.08em
    T\kern-.1667em\lower.7ex\hbox{E}\kern-.125emX}}
\usepackage{balance}
\newcommand{\add}[1]{\textcolor{black}{#1}}
\definecolor{rblue}{rgb}{0,0.5,1}
\definecolor{awesome}{rgb}{1.0, 0.13, 0.32}
\definecolor{hollywoodcerise}{rgb}{0.96, 0.0, 0.63}
\definecolor{lasallegreen}{rgb}{0.03, 0.47, 0.19}
\definecolor{hanpurple}{rgb}{0.32, 0.09, 0.98}
\definecolor{green(pigment)}{rgb}{0.0, 0.65, 0.31}
\usepackage[pagebackref=false,breaklinks=true,colorlinks,bookmarks=false]{hyperref}
\hypersetup{colorlinks=true,linkcolor={red},citecolor={hanpurple},urlcolor={magenta}}

\begin{document}
\title{CFMW: Cross-modality Fusion Mamba for Robust Object Detection under Adverse Weather}
\author{Haoyuan Li, Qi Hu, Binjia Zhou, You Yao, Jiacheng Lin,\\Kailun Yang, and Peng Chen, \textit{Member}, \textit{IEEE}
\thanks{This work was supported in part by Zhejiang Provincial Natural Science Foundation of China under Grant No. LDT23F0202 and No. LDT23F02021F02, in part by the National Natural Science Foundation of China (NSFC) under Grant No. 62473139, in part by the Hunan Provincial Research and Development Project under Grant No. 2025QK3019, and in part by the Open Research Project of the State Key Laboratory of Industrial Control Technology, China under Grant No. ICT2025B20.}%
\thanks{Haoyuan Li, Qi Hu, Binjia Zhou, and Peng Chen are with the School of Computer Science and Technology, Zhejiang University of Technology, Hangzhou, China.}
\thanks{You Yao is with the USC Viterbi School of Engineering, the University of Southern California, Los Angeles, California, United States.}
\thanks{Jiacheng Lin is with the College of Computer Science and Electronic Engineering, Hunan University, Changsha, China.}
\thanks{Kailun Yang is with the School of Robotics, Hunan University, Changsha, China.}

\thanks{Corresponding authors: Peng Chen (e-mail: chenpeng@zjut.edu.cn) and Kailun Yang (e-mail: kailun.yang@hnu.edu.cn).}
}

\markboth{IEEE Transactions on Circuits and Systems for Video Technology, July~2025}%
{CFMW}

\maketitle

\begin{abstract}
Visible-infrared image pairs provide complementary information, enhancing the reliability and robustness of object detection applications in real-world scenarios. However, most existing methods face challenges in maintaining robustness under complex weather conditions, which limits their applicability.  Meanwhile, the reliance on attention mechanisms in modality fusion introduces significant computational complexity and storage overhead, particularly when dealing with high-resolution images. To address these challenges, we propose the Cross-modality Fusion Mamba with Weather-removal (CFMW) to augment stability and cost-effectiveness under adverse weather conditions. Leveraging the proposed Perturbation-Adaptive Diffusion Model (PADM) and Cross-modality Fusion Mamba (CFM) modules, CFMW is able to reconstruct visual features affected by adverse weather, enriching the representation of image details. With efficient architecture design, CFMW is $3$ times faster than Transformer-style fusion (\textit{e.g.}, CFT). To bridge the gap in relevant datasets, we construct a new Severe Weather Visible-Infrared (SWVI) dataset, encompassing diverse adverse weather scenarios such as rain, haze, and snow. The dataset contains $64,281$ paired visible-infrared images, providing a valuable resource for future research. Extensive experiments on public datasets (\textit{i.e.}, M3FD and LLVIP) and the newly constructed SWVI dataset conclusively demonstrate that CFMW achieves state-of-the-art detection performance. Both the dataset and source code will be made publicly available at \url{https://github.com/lhy-zjut/CFMW}.

\end{abstract}


\begin{figure}[!htbp]
     \centering
    	\includegraphics[width=\linewidth]{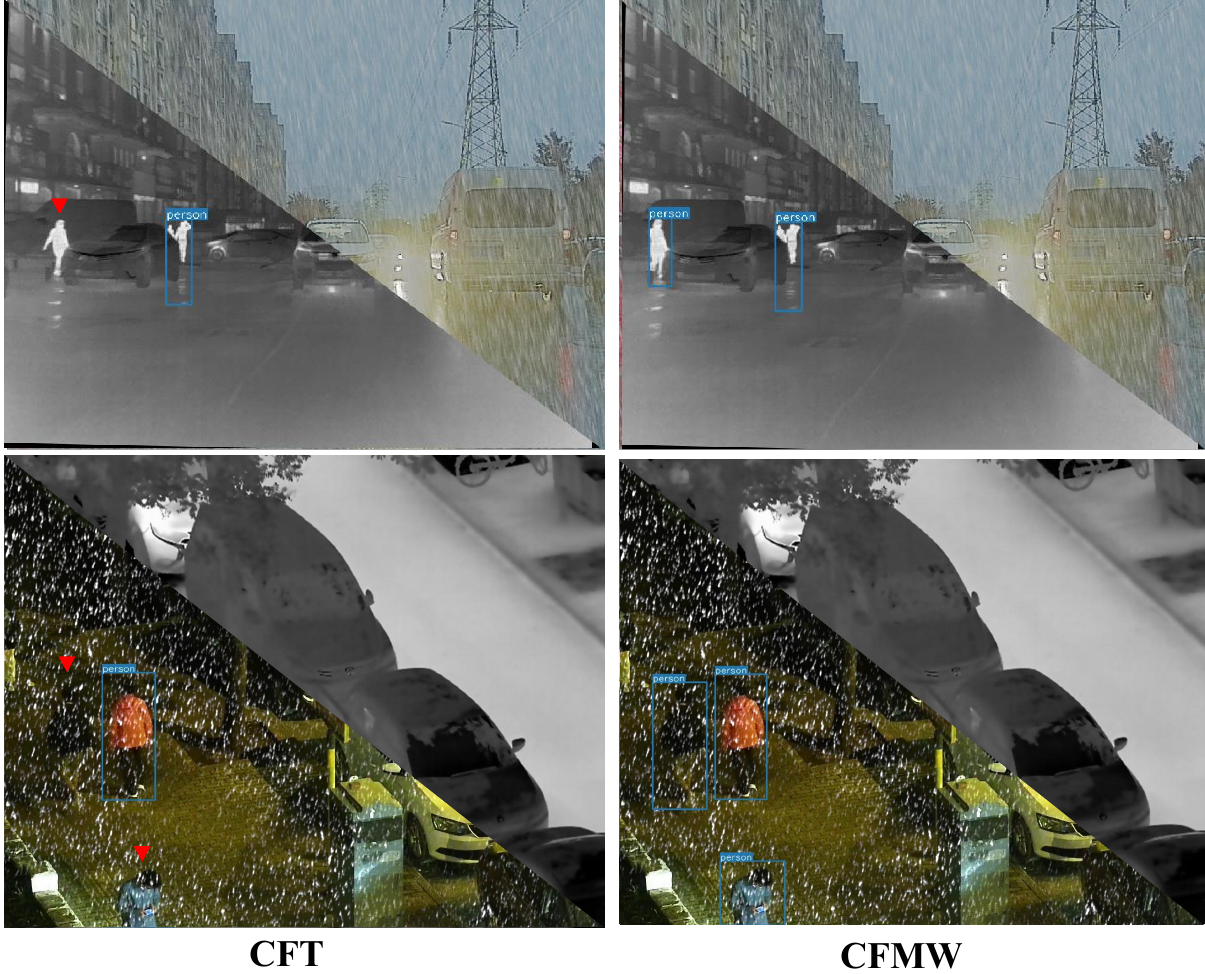}
    		\centering		
     \caption{CFMW can achieve better cross-modality object detection under adverse weather conditions than CFT \cite{qingyun2021cross}. The inverted triangle indicates the FNs.}
     \label{fig:teaser} 
\end{figure}

\section{Introduction}
    \noindent \lettrine[lines=2]{O}{bject} detection methods have experienced significant performance improvements with the rapid advancement of deep learning and have been widely used in various fields, \textit{e.g.}, autonomous driving~\cite{wang2023multi}, robotics~\cite{xu2022infrared}, tracking~\cite{li2018learning}, and person re-identification~\cite{liu2022revisiting,wu2024asymmetric,ling2023bridge}. However, it is difficult for an algorithm to use only visible-band sensor data to achieve high accuracy under occlusion, poor lighting, and adverse weather conditions~\cite{qingyun2021cross}. Unlike ordinary cameras based on visible light imaging mechanisms, infrared light is invisible to the naked eye. Infrared sensors can obtain temperature information in the scene without being restricted by natural light conditions~\cite{li2023multiscale}. Therefore, the acquired thermal infrared images can reveal the contour features of the target object in such cases. However, using infrared images alone will lead to loss of texture information, such as the color of the objects~\cite{cong2022does}. Benefiting from advanced feature extraction and fusion strategies, cross-modality fusion object detection combines the rich texture information of visible features and the contour information of infrared features, which achieves a more robust detection effect than a single modality.
    
    Existing cross-modality object detection methods can be mainly divided into traditional and deep-learning-based methods. Krotosky and Trivedi~\cite{krotosky2007color} introduced a method that extracts features from visible and infrared images by Histogram of Oriented Gradient (HOG) and then inputs the cascaded fused features into Support Vector Machines (SVM) to obtain detection results. However, methods based on hand-designed operators often struggle to achieve optimal results. Therefore, current deep-learning-based methods get rid of the manual design part and automatically extract the best features by learning the optimal parameters through neural networks. Those methods could be categorized into three strategies: pixel-level fusion~\cite{liu2023lightweight,li2023text}, feature-level fusion~\cite{li2025contourlet,chen2022multimodal}, and decision-level fusion~\cite{yao2019object}. \add{Pixel-level fusion performs cross-modal information integration prior to deep feature extraction, typically relying on a single-decoder architecture. Decision-level fusion combines the detection results from each stream of the dual-stream network only at the final stage, relying heavily on the individual feature extraction capabilities of each detection network. In contrast, feature-level fusion usually adopts a dual-stream network design to separately extract both shallow and deep features from RGB and thermal modalities, which are then fused through element-wise operations, enabling the model to capture richer and more complementary cross-modality representations~\cite{qingyun2021cross}.} Benefiting from advanced feature extraction and fusion strategies, cross-modality fusion methods (\textit{e.g.}, CFT~\cite{qingyun2021cross}, GAFF~\cite{ZHANG2021GuidedAF}, CFR\_3~\cite{zhang2020multispectral}) achieve high accuracy. Towards this end, we propose a novel framework named \textbf{C}ross-modality \textbf{F}usion \textbf{M}amba with \textbf{W}eather-removal (CFMW), as well as construct a new dataset, named \textbf{S}evere \textbf{W}eather \textbf{V}isible-\textbf{I}nfrared (SWVI) Dataset.
    
    In practical applications, object detection methods face challenging weather conditions such as rain, haze, and snow. As shown in Fig.~\ref{fig:teaser}, the performance of current methods is often challenged by adverse weather conditions, which impact the visibility and quality of visible images. \add{Rain and snow introduce streak-like noise patterns, whereas fog leads to reduced contrast and color degradation, both of which significantly increase the challenge of accurately identifying target objects. At the same time, given the high resolution of images in existing visible-infrared datasets, Transformer-based fusion methods that rely on attention mechanisms to model cross-modality feature similarity incur $O(n^2)$ complexity due to the computation of pairwise attention matrices. This leads to substantial memory overhead and slow inference, limiting their scalability. Recent SSMs, such as Mamba models~\cite{gu2023mamba,dao2024transformers}, leverage an input-dependent selection mechanism to address the limitations of fixed parameterization, whose similarity incurs $O(n)$ complexity. Towards this end, we propose a novel framework named \textbf{C}ross-modality \textbf{F}usion \textbf{M}amba with \textbf{W}eather-removal (CFMW), as well as construct a new dataset, named \textbf{S}evere \textbf{W}eather \textbf{V}isible-\textbf{I}nfrared (SWVI) Dataset. } 

    Motivated by the failure cases highlighted in Fig.~\ref{fig:teaser}, we introduce CFMW for cross-modality object detection under adverse weather conditions. Our CFMW leverages a \textbf{P}erturbation-\textbf{A}daptive \textbf{D}iffusion \textbf{M}odel (PADM) and \textbf{C}ross-modality \textbf{F}usion \textbf{M}amba (CFM) to enhance detection accuracy amid adverse weather conditions while minimizing computational burden. Specifically, PADM is employed to restore the quality of visible images affected by adverse weather before fusion with infrared counterparts. Based on learning reversal to increase the order of noise and disrupt the process of data samples, the PADM model is advantageous in minimizing the impact of adverse weather conditions. Additionally, CFM can be integrated into the feature extraction backbone, effectively integrating global contextual information from diverse modalities. Recent research shows that Mamba~\cite{gu2023mamba} achieves higher inference speed than the equivalent-scale transformer.
    
    To facilitate research in this area, we propose a new visible-infrared dataset, named SWVI, which is designed to encompass diverse severe weather scenarios by mathematically formalizing the impact of various weather phenomena on images. Specifically, SWVI comprises $64,281$ aligned visible-infrared images, spanning $3$ weather conditions and $2$ scenes, with an even distribution across each condition and scene.
    
    Extensive experiments on both well-established and self-created datasets demonstrate that our CFMW method achieves superior detection performance compared to existing benchmarks. Specifically, we achieved about $17\%$ performance improvement compared with the current state-of-the-art image restoration methods. The proposed method achieves about $8\%$ accuracy improvement with $3$ times faster than CFT~\cite{qingyun2021cross}, a state-of-the-art cross-modality object detection method. The main contributions of this work are summarized as follows: 
    \begin{itemize}
        \item We introduce a novel task focusing on visible-infrared object detection under adverse weather conditions and develop a new dataset called the Severe Weather Visible-Infrared Dataset (SWVI), which simulates real-world conditions. SWVI comprises $64,281$ paired visible-infrared images and labels, encompassing weather conditions such as rain, haze, and snow.
        \item We propose a novel approach, Cross-modality Fusion Mamba with Weather-removal (CFMW), for visible-infrared object detection under adverse weather.
        \item We introduce a novel Perturbation-Adaptive Diffusion Model (PADM) and Cross-modality Fusion Mamba (CFM) modules to tackle image de-weathering and visible-infrared object detection tasks simultaneously.
        \item Extensive experiments demonstrate proposed CFMW achieves state-of-the-art performance in multiple datasets.
    \end{itemize}

\begin{figure*}[!t]
    \centering        
    \includegraphics[width=\linewidth]{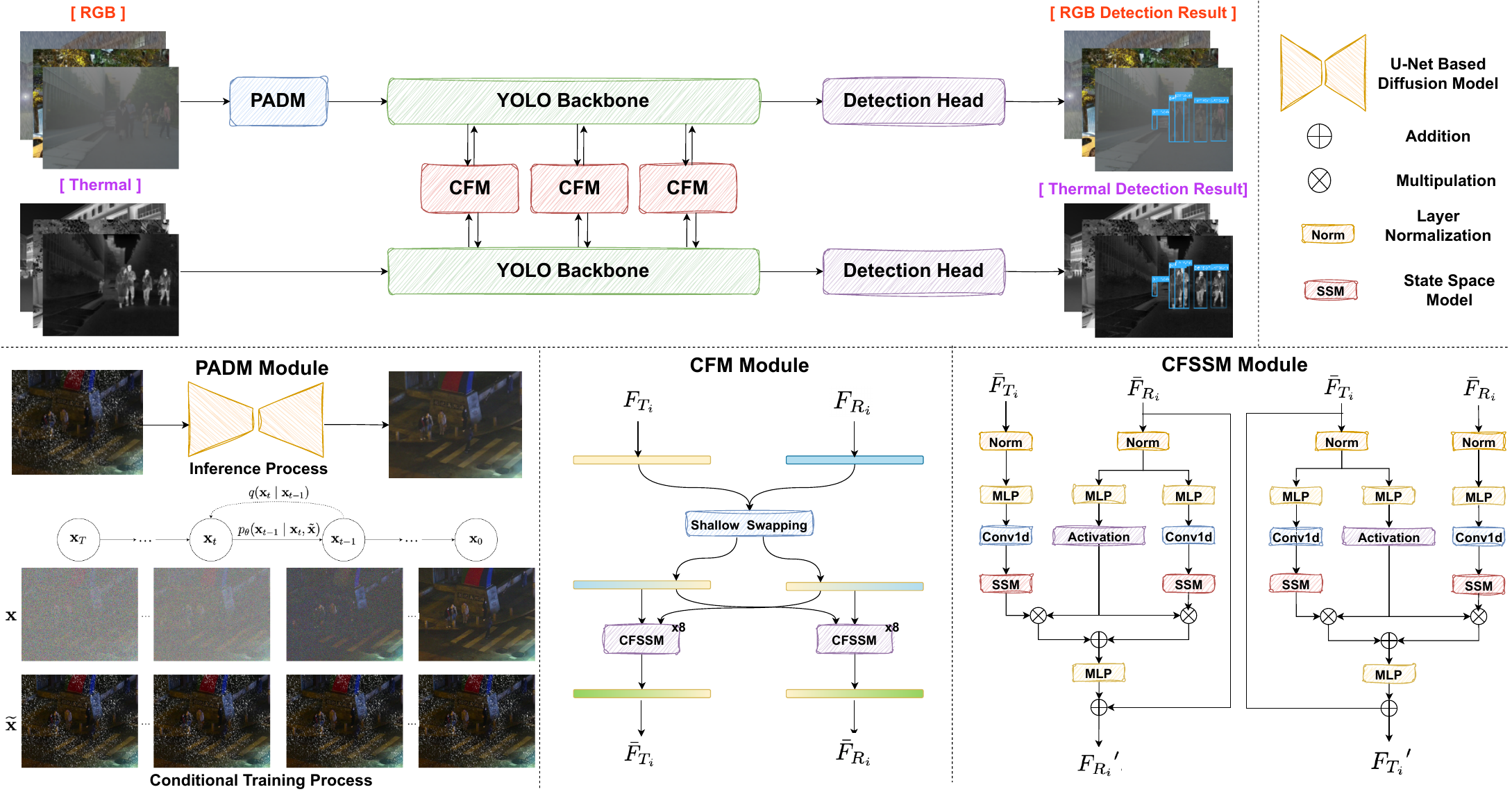}
    \caption{Framework of Cross-Modality Fusion Mamba. \add{The core pipeline of CFMW primarily consists of a YOLO detection network as backbone, a PADM module, and three CFM modules. Notice that $\bigoplus$ represents element-wise add, $\bigotimes$ represents element-wise multiply. In PADM module, $\mathbf{x}$ represents noising image, $\mathbf{\widetilde{x}}$ represents conditional weather-influenced image, and $\mathbf{x}_t$ represents the noising image during $t$-th diffusion step (here $t\in[0,T]$). As illustrated in the lower-left portion of the figure, the PADM model performs a $T$-step denoising process during inference in total to recover the original image features unaffected by adverse weather conditions. In CFM and CFSSM modules, $F_{R_i}$ and $F_{T_i}$ denote the image features extracted from the $i$-th layer of the backbone network for the RGB and thermal modalities, respectively. $\bar{F}_{R_i}$ and $\bar{F}_{T_i}$ represent feature processed by CFM module, and ${F_{R_i}}'$ and ${F_{T_i}}'$ represent feature processed by CFSSM module.}}
    \label{fig2}
\end{figure*}

\section{Related Work}
\noindent In this section, we briefly summarize the recent development of cross-modality object detection. We also briefly review previous related works about state space models and multi-weather image restoration.
 
\noindent\textbf{Cross-modality Object Detection.}
Existing cross-modality object detection methods can be divided into three categories: feature-level, pixel-level, and decision-level fusion, distinguished through feature fusion methods and timing. Recently, dual-stream object detection models based on convolutional neural networks have made significant progress in improving recognition performance~\cite{zeng2024mmi,chen2021multimodal,qingyun2021cross,zhang2023differential}, while pixel-level fusion methods have also achieved promising results~\cite{zhao2023cddfuse,cheng2023mufusion,li2020spectrum,bai2024ibfusion}. 
Other works employing methods such as Generative Adversarial Network (GAN) to effective integration also have achieved good results~\cite{gao2022dcdr,zhao2023interactive,zhao2023cddfuse,li2023mrfddgan,ma2020ganmcc}. These approaches can be integrated into downstream tasks such as object detection. 
Traditional convolutional neural networks have limited receptive fields that the information is only integrated into a local area when using the convolutional operator, whereas the self-attention mechanism of transformers enables the learning of long-range dependencies~\cite{vaswani2017attention}. 
Thus, a transformer-based method, named Cross-Modality Fusion Transformer (CFT)~\cite{qingyun2021cross}, was presented and achieved state-of-the-art detection performance. Differing from these works, we introduce Mamba into cross-modality object detection to learn long-range dependencies, achieving high accuracy and low computation overhead.\\
\noindent\textbf{State Space Model.}
The concept of the state space model was initially introduced in the \textbf{S}tructured \textbf{S}tate \textbf{S}pace \textbf{S}equence models (S4)~\cite{gu2021efficiently}. Compared with traditional convolutional neural networks and Transformer-style methods, the S4 model presents a distinctive architecture capable of effectively modeling global information. Based on S4, the S5 model~\cite{smith2022simplified} reduces complexity to a linear level, with H3~\cite{mehta2022long} introducing it into language model tasks. Mamba~\cite{gu2023mamba} introduced an input-activated mechanism to enhance the state space model, achieving higher inference speed and overall metrics compared with equivalent-scale transformers. With the introduction of Vision Mamba~\cite{zhu2024vision} and VMamba~\cite{liu2024vmamba}, the application of the state space model has been extended into visual tasks. Currently, existing research does not consider effectively generalizing the state space model to cross-modality object detection.\\
\noindent\textbf{Multi-Weather Image Restoration.} 
Recently, some attempts have been made to unify multiple recovery tasks in a single deep learning framework, including generating modeling solutions to recover superimposed noise types~\cite{feng2021deep}, recovering superimposed noise or weather damage with unknown test time, or especially unfavorable multi-weather image fading~\cite{valanarasu2022transweather,chen2022learning,li2022all}. 
All-in-One~\cite{li2020all} unified a weather restoration method with a multi-encoder and decoder architecture. \add{GridFormer~\cite{wang2024gridformer} introduces a residual dense transformer with a grid structure, utilizing an enhanced attention mechanism and residual dense transformer blocks for multi-weather restoration. MB-TaylorFormer V2~\cite{jin2025mb} proposes an improved multi-branch linear transformer expanded by the Taylor formula, capable of concurrently processing coarse-to-fine features and capturing long-distance pixel interactions with limited computational cost. ESTINet~\cite{zhang2022enhanced} presents an end-to-end video deraining framework that boosts performance by capturing spatial features and temporal correlations between consecutive frames. Dual Attention-in-Attention Model~\cite{zhang2021dual} develops a model that includes two dual-attention modules to address both rain streaks and raindrops simultaneously.} 
It is worth noting that diffusion-based conditional generative models have shown state-of-the-art performance in various tasks such as class-conditional data synthesis with classifier guidance~\cite{dhariwal2021diffusion}, image super-resolution~\cite{ho2022cascaded}, image deblurring~\cite{whang2022deblurring}. Denoising Diffusion Restoration Models (DDRM)~\cite{kawar2022denoising} were proposed for general linear inverse image restoration problems. \add{WeatherDiff~\cite{ozdenizci2023} is the first to introduce conditional denoising diffusion models into multi-weather restoration and has achieved impressive results.} Unlike existing works, we expand the multi-weather restoration to enhance the model's robustness to adverse weather conditions. \add{By leveraging Mamba~\cite{gu2023mamba} blocks to capture the contextual relationships within image features, our proposed PADM effectively restores details in images that are degraded by weather-induced noise.}

\section{method}
\subsection{Overview}
\label{overview}
\noindent To achieve the purpose of detecting objects efficiently, we construct a framework named CFMW, as illustrated in Fig.~\ref{fig2}. The whole framework consists of four parts: PADM module, YOLO network backbone, CFM blocks, and detection head. In detail, the CFM module achieves efficient cross-modality feature fusion, while the PADM enhances the robustness of the framework under adverse weather conditions.

\begin{figure}[!t]
        \centering
        \includegraphics[width=\linewidth]{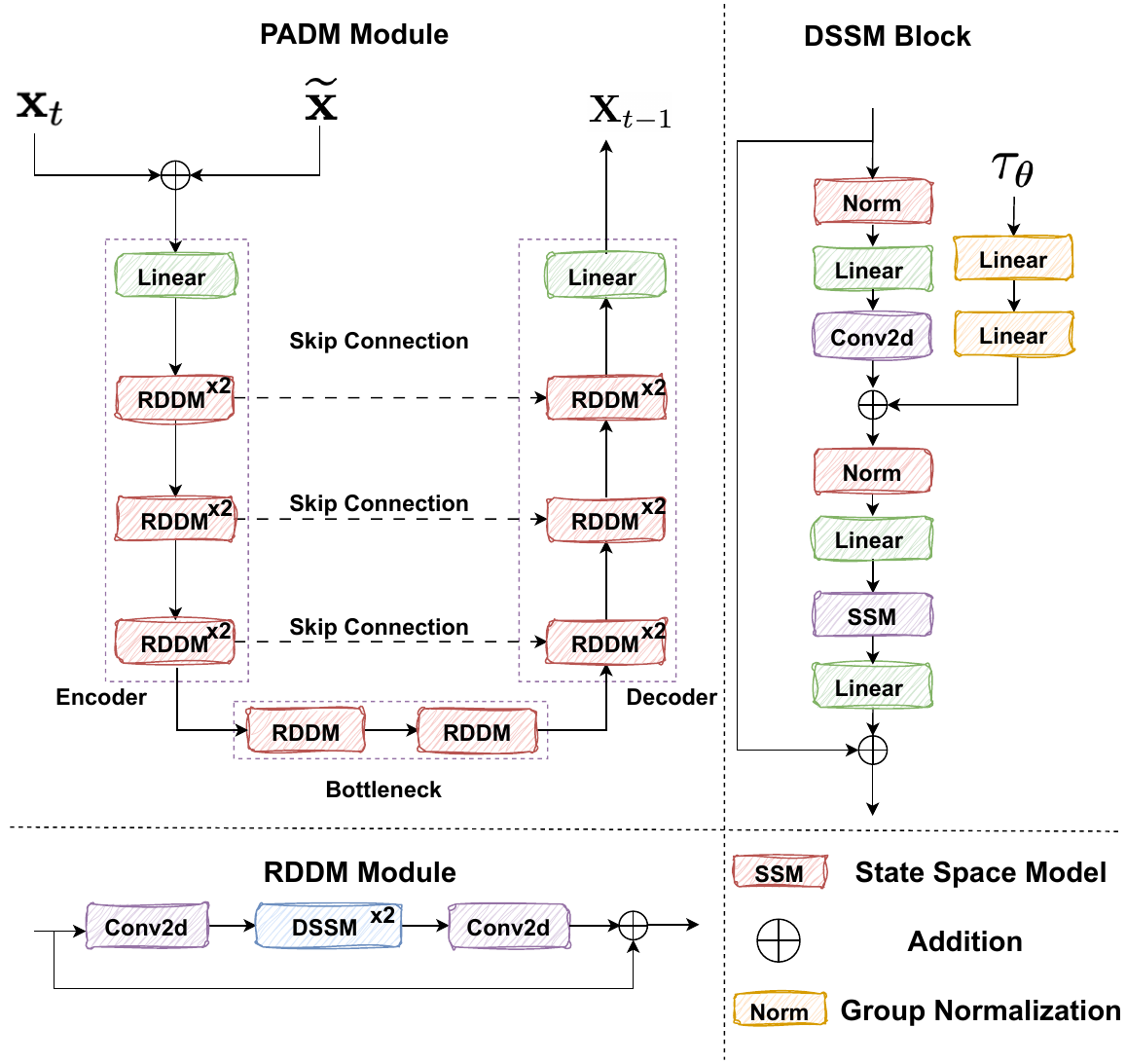}
    \caption{\add{Overview of the forward diffusion and reverse denoising processes for a conditional diffusion model. Notice that $\bigoplus$ represents element-wise add, $\mathbf{x}_t$ represents the noising image during $t$-th diffusion step (here $t\in[0,T]$), $\mathbf{\widetilde{x}}$ represents conditional weather-influenced image, and $\tau_\theta$ represents the original representation of diffusion step.}}
    \label{fig3}
\end{figure}

\subsection{Perturbation-Adaptive Diffusion Model}
\label{PADM}
\noindent Denoising diffusion models~\cite{sohl2015deep,ho2020denoising} are a class of generative models, which learn a Markov chain that gradually transforms a Gaussian noise distribution into the data distribution trained by the models. The original denoising diffusion probabilistic models (DDPMs)~\cite{ho2020denoising} diffusion process (data to noise) and generative process (noise to data) are based on a Markov chain process, resulting in a large number of steps and huge time consumption. Thus, Denoising Diffusion Implicit Models (DDIMs)~\cite{song2020denoising} were presented to accelerate sampling, providing a more efficient class of iterative implicit probabilistic models. DDIMs define the generative process via a class of non-Markovian diffusion processes that lead to the same training objective as DDPMs but can produce deterministic generative processes, thus speeding up sample generation. In DDIMs, implicit sampling refers to the generation of samples from the latent space of the model in a deterministic manner. We can prove by mathematical induction that for all $t$:
\begin{equation}
    q_\lambda(\bm{\mathrm{X}}_{t-1}|\bm{\mathrm{X}}_t,\bm{\mathrm{X}}_0)=\mathcal N(\bm{\mathrm{X}}_{t-1};\bm{\widetilde{\mu}_t}(\bm{\mathrm{X}}_t,\bm{\mathrm{X}}_0),\beta_t\bm{I}),
\end{equation}
\begin{eqnarray}
    \bm{{\widetilde{\mu}}}_t=\sqrt{\bar{\alpha}_{t-1}} \mathrm{X}_0+\sqrt{1-\bar{\alpha}_{t-1}-\beta_t}\cdot\bm{\epsilon}_t,
\end{eqnarray}
\begin{equation}
\begin{aligned}
    \bm{\mathrm{X}}_{t-1}=\sqrt{\bar{\alpha}_{t-1}}\cdot(\frac{\bm{\mathrm{X}}_t-\sqrt{1-\bar{\alpha}_t}\cdot\epsilon_\theta(\bm{\mathrm{X}}_t,t)}{\sqrt{\bar{\alpha}}_t})\\
    +\sqrt{1-\bar{\alpha_{t-1}}}\cdot\epsilon_\theta(\bm{\mathrm{X}}_t,t),
\end{aligned} 
\label{eq1}
\end{equation}
where $\bm{\mathrm{X}}_t$ and $\bm{\mathrm{X}}_{t-1}$ represent the data $\bm{\mathrm{X}}_0\sim q(\bm{\mathrm{X}}_0)$ in different diffusion time steps, $\alpha_t=1-\beta_t$, $\bar{\alpha}_t=\prod\limits_{i=1}^t\alpha_i$,
and $\epsilon_\theta(\bm{\mathrm{X}}_t,t)$ can be optimized as: 
\begin{equation}
    \mathbb{E}_{\bm{\mathrm{X}}_0,t},\epsilon_t\sim \mathcal N(\bm{0,I}),[\Vert\epsilon_t-\epsilon_\theta(\sqrt{\bar{\alpha}_t}\bm{\mathrm{X}}_0+\sqrt{1-\bar{\alpha}_t}\epsilon_t,t\Vert^2].
\end{equation}

Conditional diffusion models have shown state-of-the-art image-conditional data synthesis and editing capabilities \cite{dhariwal2021diffusion,Choi2021ILVRCM,ozdenizci2023}. 
The core idea is to learn a conditional reverse process without changing the diffusion process. 
Our proposed PADM is a conditional diffusion model, adding reference images (clear images) in the process of sampling to guide the reconstructed image to be similar to the reference images.

Specifically, as shown in Fig.~\ref{fig3}, we introduce a new parameter $\widetilde{\bm{\mathrm{X}}}$, which represents the weather-degraded observation. 
A Markov chain is defined as a diffusion process, and Gaussian noise is gradually added to simulate the gradual degradation of data samples until reaching time point $\mathrm{T}$. We ground our model hyperparameters via a U-Net architecture based on WideResNet~\cite{Zagoruyko2016WideRN}. 
For the input images' conditional reflection, we connect patch $\bm{\mathrm{X}}_T$ and $\widetilde {\bm{\mathrm{X}}}$, to obtain the six-dimensional input image channel.
Conditioning the reverse process on $\widetilde{\mathrm{x}}$ can maintain its compatibility with implicit sampling, so we could expand Eq.~(\ref{eq1}) as:
\begin{equation}
\begin{aligned}
    \bm{\mathrm{X}}_{t-1}=\sqrt{\bar{\alpha}_{t-1}}\cdot(\frac{\bm{\mathrm{X}}_t-\sqrt{1-\bar{\alpha}_t}\cdot\epsilon_\theta(\bm{\mathrm{X}}_t,\widetilde{\bm{\mathrm{X}}},t)}{\sqrt{\bar{\alpha}}_t})\\ +\sqrt{1-\bar{\alpha_{t-1}}}\cdot\epsilon_\theta(\bm{\mathrm{X}}_t,\widetilde{\bm{\mathrm{X}}},t).
\end{aligned}
\end{equation}
The sampling process starts from $\bm{\mathrm{X}}_T\sim \mathcal{N}(\bm{0,I})$,  following a deterministic reverse path towards $\bm{\mathrm{X}}_0$ with fidelity.

\subsection{Cross-modality Fusion Mamba}
\label{CFM}

\add{The most straightforward way is to utilize concatenation, element-wise addition, element-wise average/maximum, and element-wise cross product to merge feature maps of visible and infrared modalities directly. Fang~\textit{et al.}~\cite{qingyun2021cross} proposed a Transformer-based scheme to fuse intra-modal and inter-modal information for multispectral. However, due to the high computational overhead introduced by the multi-head attention mechanism, such modality fusion methods are not well-suited for high-resolution scenarios. Advanced State Space Model (SSM), or Mamba~\cite{gu2023mamba}, is more efficient and faster than Transformer-style methods when processing long sequences thanks to its linear complexity and hardware adaptability. Therefore, we designed the CFM module with the goal of leveraging the linear computational complexity of Mamba to handle high-resolution detection tasks more efficiently.} 
The details of the CFM module are shown in Fig.~\ref{fig3}.

S4~\cite{gu2021efficiently} and Mamba~\cite{gu2023mamba} are inspired by the continuous system, mapping a 1-D function or sequence $x(t)\in \mathbb{R}^N\to y(t)$ through a hidden state $h(t)\in \mathbb{R}^N$. 
This system uses $\bm{A}\in \mathbb{R}^{N\times N}$ as the evolution parameter and $\bm{B}\in \mathbb{R}^{N\times 1}, \bm{C}\in \mathbb{R}^{1\times N}$ as the projection parameters, so that $y(t)$ could evolve as follows: 
\begin{equation}
\begin{aligned}
    &{h}'(t)=\bm{A}h(t)+\bm{B}x(t), \\
    &y(t)=\bm{C}{h}'(t).
\end{aligned}
\label{eq3}
\end{equation}

\begin{figure}[!t]
        \centering
        \includegraphics[width=\linewidth]{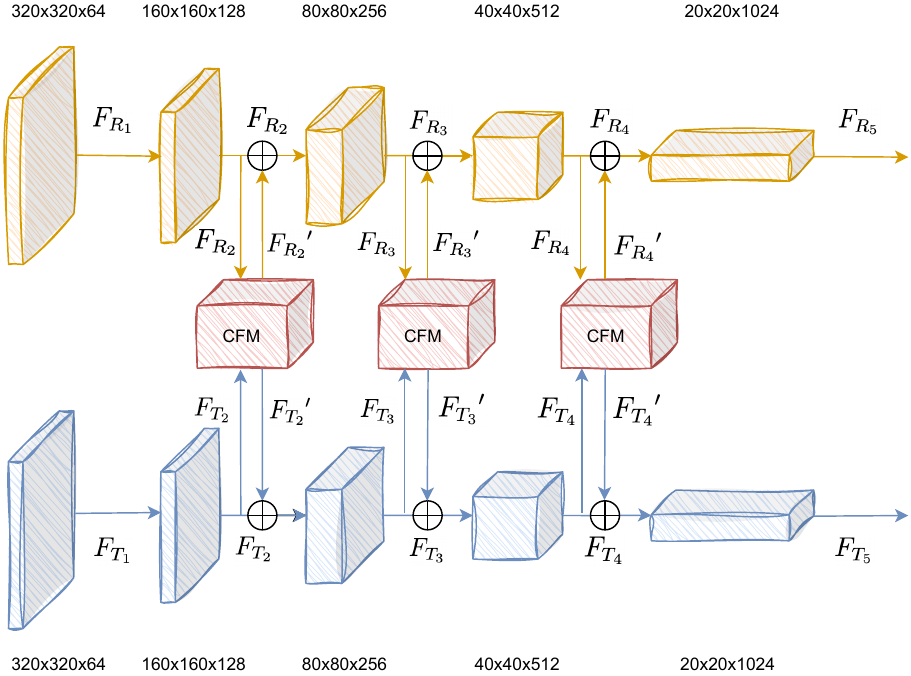}
    \caption{
    Details of the YOLO Backbone and CFM block. \add{Notice that $\bigoplus$ represents element-wise add. $F_{R_i}$ and $F_{T_i}$ denote the image features extracted from the $i$-th layer of the backbone network for the RGB and thermal modalities, respectively. The orange and blue connection lines represent features from RGB and thermal images, and the purple lines represent the concatenated features of both modalities.}}
    \label{fig_CFM}
\end{figure}  

Notice that S4 and Mamba are the discrete versions of the continuous system, including a timescale parameter $\Delta$ to transform the continuous parameters $A,B$ to discrete parameters $\bar{\bm{A}},\bar{\bm{B}}$ as follows:
\begin{equation}
\begin{aligned}
    &\bar{\bm{A}}=\mathrm{exp}(\Delta \bm{A}), \\
    &\bar{\bm{B}}=(\Delta \bm{A})^{-1} (\mathrm{exp}(\Delta \bm{A})- \bm{I})\cdot\Delta \bm{B}.
\end{aligned} 
\end{equation}
After that, Eq.~(\ref{eq3}) could be rewritten as:
\begin{equation}
\begin{aligned}
    &h_t=\bar{\bm{A}}h_{t-1}+\bar{\bm{B}}\bm{X_t},\\
    &y_T=\bm{C}h_t.
\end{aligned}
\end{equation}
Finally, the models compute output through a global convolutional layer as follows:
\begin{equation}
\begin{aligned}
    &\bar{\bm{K}}=\bm{C\bar{B}},\bm{C\bar{A\bar{B}}},...,\bm{C\bar{A}^{L-1}\bar{B}},\\
    &y=x*\bar{\bm{K}},
\end{aligned}
\end{equation}
where $L$ is the length of the input sequence $x$, and $\bar{\bm{K}}\in \mathbb{R}^M$ is a structured convolutional kernel.

The standard Mamba is designed for the 1-D sequence. As shown in Vision Mamba (Vim)~\cite{zhu2024vision},  2-D multispectral images $t\in \mathbb{R}^{H\times W\times C}$ could be transformed into the flattened 2-D patches $\bm{X}_p\in \mathbb{R}^{J\times(P^2\times C)}$, where $(H,W)$ represents the size of input images, $C$ is the channels, and $P$ is the size of image patches. 
Similarly, we linearly project the $x_p$ to the vector with size $D$ and add position embeddings $\bm{E}_{pos}\in \mathbb{R}^{(J+1)\times D}$ as follows:
\begin{equation}
    \bm{T}_0 = [t_{cls}; t_p^1\bm{W}; t_p^2\bm{W}; ...; t_p^J\bm{W}]+\bm{E}_{pos},
\end{equation}
where $t_P^j$ is the $j$-th path of $t$, $\bm{W}\in \mathbb{R}^{(P^2\times C)\times D}$ is the learnable projection matrix. 

Here are more details of the proposed CFM. As mentioned in the introduction section, the RGB modality and the Thermal modality show different features under different lighting and weather conditions, which are complementary and redundant. Therefore, we aim to design a block to suppress redundant features and fuse complementary information to efficiently harvest essential cross-modal cues for object detection against adverse weather conditions. Motivated by the concept of Cross-Attention~\cite{chen2021crossvit}, we introduce a new cross-modality Mamba block to fuse features from different modalities. As shown in Fig.~\ref{fig2}, to encourage feature interaction between RGB and Thermal modalities, we first use a shallow swapping block, which incorporates information from different channels and enhances cross-modality correlations. Given RGB features $\bm{F}_{R_{i}}\in \mathbb{R}^{B\times N \times C}$ and Thermal features $\bm{F}_{T_{i}}\in \mathbb{R}^{B\times N \times C}$, the first half of channels from $F_{R_{i}}$ ($\bm{F}_{R_i}^{\text{front}}$) will be concatenated with the latter half of $\bm{F}_{T_{i}}$ ($\bm{F}_{T_i}^{\text{back}}$). The obtained features are added to $\bm{F}_{R_{i}}$, creating a new feature $\bar{\bm{F}}_{R_i}\in \mathbb{R}^{B\times N \times C}$. Meanwhile, the first half of $\bm{F}_{T_{i}}$ ($\bm{F}_{T_i}^{\text{front}}$) is concatenated with the latter half of $\bm{F}_{R_{i}}$ ($\bm{F}_{R_i}^{\text{back}}$). The obtained features are added to $\bm{F}_{T_{i}}$, creating a new feature $\bar{\bm{F}}_{T_i}\in \mathbb{R}^{B\times N \times C}$. This process can be expressed by the following formula:
\begin{equation}
\begin{aligned}
    &\bm{F}_{R_i^{\text{front}}} = \bm{F}_{R_i}[:, :, :C/2],\\ 
    &\bm{F}_{T_i}^{\text{back}} = \bm{F}_{T_i}[:, :, C/2:];
\end{aligned}
\end{equation}
\begin{equation}
\begin{aligned}
    &\bm{F}_{R_i} = \text{Concat}(\bm{F}_{R_i}^{\text{front}},\bm{F}_{T_i}^{\text{back}}),\\
    &\bm{F}_{T_i} = \text{Concat}(\bm{F}_{T_i}^{\text{front}}, \bm{F}_{R_i}^{\text{back}}).
\end{aligned}
\end{equation}

Subsequently, we project the features: $\bar{\bm{F}}_{R_i}$ and $\bar{\bm{F}}_{T_i}$ into the shared space during the feature fusion process, using the gating mechanism to encourage complementary feature learning while restraining redundant features. As shown in Fig.~\ref{fig2}, we first normalize every token sequence in $\bar{\bm{F}}_{R_i}$ and $\bar{\bm{F}}_{T_i}$ with Norm block, which helps to improve the convergence speed and performance of the model. 
Then, we project the input sequence through a 3-layer MLP and apply SiLU as the activation function. 
After that, we apply 2D-Selective-Scan method proposed by VMamba~\cite{liu2024vmamba}:
\begin{equation}
\begin{aligned}
    &y_R = \mathrm{SS2D}  {({\bar{\bm{F}}_{R_i}})},\\
    &y_T = \mathrm{SS2D}(\bar{\bm{F}}_{T_i}).
\end{aligned}
\end{equation}
Then we apply the gating operation, followed by a residual connection. 
\begin{equation}
\begin{aligned}
     &\bm{Z}_T = \mathrm{MLP}(\bar{\bm{F}}_{T_{i}}),\\
     &\bm{Z}_R = \mathrm{MLP}(\bar{\bm{F}}_{R_{i}});
\end{aligned}
\end{equation}
\begin{equation}
\begin{aligned}
    {y_R}'= y_R\odot \mathrm{SiLU}(\bm{Z_R}),\\
    {y_T}'= y_T\odot \mathrm{SiLU}(\bm{Z_T}); 
\end{aligned}
\end{equation}
\begin{equation}
\begin{aligned}
    \hat{\bm{F}}_{T_i}&= \mathrm{MLP} {({y_R}'+{y_T}')+\bar{\bm{F}}_{R_i})},\\
    \hat{\bm{F}}_{R_i}&= \mathrm{MLP} {({y_R}'+{y_T}')+\bar{\bm{F}}_{T_i})}.
\end{aligned}
\end{equation}
\begin{equation}
    \begin{aligned}
        &{\bm{F}_{T_i}}'=\bar{\bm{F}}_{T_i}+\hat{\bm{F}}_{T_i},\\
        &{\bm{F}_{R_i}}'=\bar{\bm{F}}_{R_i}+\hat{\bm{F}}_{R_i},
    \end{aligned}
\end{equation}
where $\odot$ represents element-wise multiplication.

As shown in Fig.~\ref{fig_CFM}, after multiple layers of convolutional, the cross-modality features are fused in three different dimensions through CFM and finally input together into the detection head. Those fused features contain both low-dimensional and high-dimensional features, which enhances the network's ability to perceive the overall image and capture image details.
\begin{equation}
    \begin{aligned}
        &\bm{F}_{T_i}=\bm{F}_{T_i}+{{\bm{F}_{T_i}}}',\\
        &\bm{F}_{R_i}=\bm{F}_{R_i}+{{\bm{F}_{R_i}}}',
    \end{aligned}
\end{equation}
where $i\in {2,3,4}$.

Different from CFT~\cite{qingyun2021cross}, our fusion block improves computational efficiency while inheriting the components of global receptive field and dynamic weight. Comparing the State Space Model (SSM) in our CFM block with the self-attention mechanism of transformers in CFT~\cite{qingyun2021cross}, both of them play an important role in providing global context adaptively, but self-attention is quadratic to sequence length while SSM is linear to sequence length~\cite{zhu2024vision}. 
To achieve lower memory usage when dealing with long-sequence works, CFM chooses the recomputation method as the same as Mamba, including recomputing intermediate activations such as the output of activation functions and convolutions which take a lot of GPU memory but are fast for recomputation. 
Meanwhile, the time complexity of the Transformer’s attention mechanism is $O(n^2)$, whereas Mamba’s time complexity is $O(n)$ ($n$ represents the sequence length). 

\begin{table*}[!t]
\caption{Comparisons of the SWVI benchmark with existing visible-infrared datasets. Here \textcolor{black}{\ding{51}} means available while \textcolor{black}{\ding{55}} means unavailable.}
    \setlength{\tabcolsep}{3pt} 
    \renewcommand{\arraystretch}{1.1} 
    \resizebox{\linewidth}{!}{
    \begin{tabular}{cccccccccc}
    \hline
    \multicolumn{1}{c}{\multirow{2}{*}{Dataset}} & \multirow{2}{*}{Year} & \multicolumn{1}{c}{\multirow{2}{*}{Resolution}} & \multirow{2}{*}{Publication} & \multicolumn{3}{c}{Scene} & \multirow{2}{*}{Camera Angle} & \multirow{2}{*}{\#Image} & \multirow{2}{*}{Annotation} \\ \cline{5-7} 
    \multicolumn{1}{c}{} & & \multicolumn{1}{c}{} & & \multicolumn{1}{c}{Daylight} & \multicolumn{1}{c}{Night} & \multicolumn{1}{c}{Weather} & & & \\ \hline
    KAIST~\cite{hwang2015multispectral} & 2015 & $640\times512$ & CVPR & \textcolor{black}{62.5\%} & \textcolor{black}{37.5\%} & \textcolor{black}{\ding{55}} & Horizontal & 95328 & \textcolor{black}{\ding{51}} \\
    VEDAI~\cite{razakarivony2016vehicle} & 2016 & $512^2\&1024^2$ & Vis. Commun. Image Represent. & \textcolor{black}{62.5\%} & \textcolor{black}{37.5\%} & \textcolor{black}{\ding{55}} & Remote sensing & 3364 & \textcolor{black}{\ding{51}} \\
    FLIR~\cite{FLIR} & 2018 & $640\times512$ & - & \textcolor{black}{60.2\%} & \textcolor{black}{39.8\%} & \textcolor{black}{\ding{55}} & Driving & 14452 & \textcolor{black}{\ding{51}} \\
    RoadScene~\cite{xu2020aaai} & 2020 & $640\times512$ & AAAI & \textcolor{black}{71.3\%} & \textcolor{black}{28.7\%} & \textcolor{black}{\ding{55}} & Driving & 442 & \textcolor{black}{\ding{51}} \\
    Freiburg Thermal~\cite{vertens20bridging} & 2020 & $640\times512$ & IROS & \textcolor{black}{58.3\%} & \textcolor{black}{41.7\%} & \textcolor{black}{\ding{55}} & Driving & 20000 & \textcolor{black}{\ding{55}} \\
    LLVIP~\cite{jia2021llvip} & 2021 & \textbf{$1280\times1024$} & ICCV & \textcolor{black}{7.6\%} & \textcolor{black}{92.4\%} & \textcolor{black}{\ding{55}} & Surveillance & 30976 & \textcolor{black}{\ding{51}} \\
    MSRS~\cite{tang2022piafusion} & 2022 & $640\times480$ & Inform. Fusion & \textcolor{black}{52.2\%} & \textcolor{black}{47.8\%} & \textcolor{black}{\ding{55}} & Horizontal & 3136 & \textcolor{black}{\ding{51}} \\
    M3FD~\cite{liu2022target} & 2022 & $1024\times768$ & CVPR & \textcolor{black}{68.9\%} & \textcolor{black}{31.1\%} & \add{12.4\%} & Horizontal & 8400 & \textcolor{black}{\ding{51}} \\
    \rowcolor{gray!15}\textbf{SWVI}& 2025 & \bm{$1280\times1024$} & \textbf{Proposed} & \textcolor{black}{63.3\%} & \textcolor{black}{36.7\%} & \textbf{100}\% & \textbf{Multiple angle} & \textbf{64281}& \textcolor{black}{\ding{51}} \\ \hline
    \label{Dataset}
    \end{tabular}
    }
\end{table*}

\subsection{Loss Functions}
\label{Loss function}
\noindent We carefully design the training loss functions to produce enhanced results with minimum blurriness and the closest details to ground-truth images and to extract the differences between RGB and thermal modalities. 

\noindent\textbf{Loss function for PADM.} 
For training PADM, the goal of the loss function in this stage is to maximize the data log-likelihood $log_{p_\theta(\bm{X_0})}$. Since maximizing this target directly is very challenging, we use variational inference to approximate this target. Variational inference approximates the true posterior distribution $p_\theta(\bm{X_0}:T)$ by introducing a variational distribution $q(X_1\colon T|\bm{X_0})$ and then minimizing the difference between these two distributions. Here, we use Kullback-Leibler (KL) divergence to measure the difference between two probability distributions. During training PADM, specifically, for each time step $t$, we have:
{
\begin{equation}
\begin{aligned}
    &\mathcal{L}_\theta=\mathbb{E}_q[log_{p_\theta}(\bm{X_0}|\bm{X_t})]-\\&\mathbb{E}_{q(\bm{\bm{X}_{t-1}}|\bm{X_t})}[D_{KL}(q(\bm{X_{t-1}}|\bm{X_t},\bm{X_0}))|p_\theta(\bm{\bm{X}_{t-1}}|\bm{X_t})],
\end{aligned}
\end{equation}
}
where the first term is the expected value of $log_{p_\theta}(\bm{X_0}|\bm{X_t})$  under the variational distribution $q(\bm{X_t})$, and the second term is the expected value of the Kullback-Leibler divergence between $q(\bm{X_{t-1}}|\bm{X_t})$ and $p_\theta(\bm{X_{t-1}}|\bm{X_t})$. Summing up the variational bounds for all time steps, we obtain the variational bound for the entire diffusion process:
{
\begin{equation}
\begin{aligned}
    &\mathcal{L}_\theta = \sum_{t=1}^T \mathbb{E}_q[log_{p_\theta}{(\bm{X}_0 | \bm{X}_t)}]- \\
    &\sum_{t=1}^{T-1} \mathbb{E}_{q(\bm{X}_{t-1}|\bm{X}_t)}\left[ D_{KL}\left(q(\bm{X}_{t-1}|\bm{X}_t,\bm{X}_0) \, \| \, p_\theta(\bm{X}_{t-1}|\bm{X}_t) \right) \right].
\end{aligned}
\end{equation}
}

\noindent\textbf{Loss function for CFM.} The overall loss function for CFM module ($\mathcal{L}_{total}$) is a sum of the bounding-box regression loss ($\mathcal{L}_{box}$), the classification loss ($\mathcal{L}_{cls}$), and the confidence loss ($\mathcal{L}_{conf}=\mathcal{L}_{noobj}+\mathcal{L}_{obj}$). We use loss weight parameters $\lambda_{box}$, $\lambda_{cls}$, and $\lambda_{conf}$ respectively to control the proportion of each loss in the total loss.
\begin{equation}
    \begin{aligned}
        &\mathcal{L}_{total}=\lambda_{box}\mathcal{L}_{box}+\lambda_{cls}\mathcal{L}_{cls}+\lambda_{conf}\mathcal{L}_{conf}\\
        &=\lambda_{box}\mathcal{L}_{box}+\lambda_{cls}\mathcal{L}_{cls}+\lambda_{conf}\mathcal{L}_{noobj}+\lambda_{conf}\mathcal{L}_{obj},
    \end{aligned}
\end{equation}
\begin{equation}
    \begin{aligned}
        &\mathcal{L}_{box}=\sum_{i=0}^{S^2}\sum_{j=0}^{N}\bm{l}_{i,j}^{obj}[1-GIoU_i],
    \end{aligned}
\end{equation}
\begin{equation}
    \begin{aligned}
        &\mathcal{L}_{cls}=\sum_{i=0}^{S^2}\sum_{j=0}^{N}\bm{l}_{i,j}^{obj}\sum_{c\in classes}p_i(c)log(\hat{p}_i(c)),
    \end{aligned}
\end{equation}
\begin{equation}
    \begin{aligned}
        &\mathcal{L}_{noobj}=\sum_{i=0}^{S^2}\sum_{j=0}^{N}\bm{l}_{i,j}^{noobj}(c_i-\hat{c}_i)^2,
    \end{aligned}
\end{equation}
\begin{equation}
    \begin{aligned}
        &\mathcal{L}_{obj}=\sum_{i=0}^{S^2}\sum_{j=0}^{N}\bm{l}_{i,j}^{obj}(c_i-\hat{c}_i)^2,
    \end{aligned}
\end{equation}
where Generalized Intersection over Union (GIoU) is employed as the predicted regression loss. $S^2$ and $N$ represent the number of image grids during prediction and the number of predicted boxes. $p(c)$ and $\hat{p}(c)$ represent the probability that the real sample is class $c$ and the probability that the network predicts the sample to be class $c$. $\bm{l}_{i,j}^{obj}$ represent whether the $j^{th}$ predicted box of the $i^{th}$ grid is a positive sample, with $\bm{l}_{i,j}^{noobj}$ represent whether the $j^{th}$ predicted box of the $i^{th}$ grid is a negative sample. 


\begin{figure}[!t]
        \includegraphics[width=\linewidth]{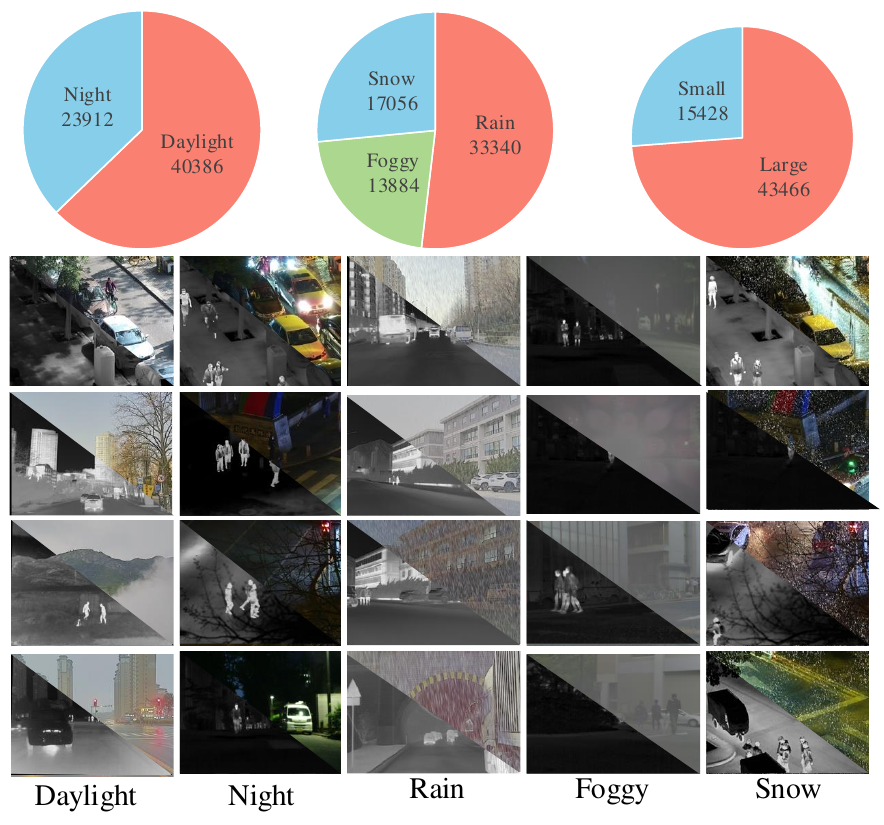}
    \caption{Overview of the established SWVI dataset. The dataset includes three weather conditions (\textit{i.e.}, Rain, Foggy, and Snow), and two scenarios (\textit{i.e.}, Daylight and Night), providing $64,281$ images in total. \add{The pie chart visualizes the proportion of images belonging to different categories in the dataset, where \textit{Large} and \textit{Small} indicate the distribution of large and small objects, respectively. Here, we define the classification criterion for object size based on whether the area of its bounding box is smaller than $2500$.}
    }
    \label{fig4}
\end{figure}

\begin{table*}[!htbp]
\centering
\renewcommand{\arraystretch}{1.2}
\caption{\add{Comparison of weather degradation models in terms of formulation and visual effects.}}
\setlength{\tabcolsep}{10pt}
\resizebox{\linewidth}{!}{
\begin{tabular}{lccc}
\toprule
\textbf{Weather} & \textbf{Modeling Idea} & \textbf{Simulation Method} & \textbf{Visual Effect} \\
\midrule
Rain & Mask + Synthesized streak & Linear blending & Fine rain marks, random rain streaks \\
Snow & Mask + Snow image & Linear blending & Local occlusion, brightness increase \\
Fog  & Atmospheric scattering model & Exponential decay + airlight addition & Overall blurring, bright distant area \\
\bottomrule
\end{tabular}}
\label{SWVI_construct}
\end{table*}

\section{Experiment}
\subsection{SWVI Benchmark}
\noindent \textbf{Dataset}. As shown in Fig.~\ref{fig4}, we established the benchmark, SWVI, which is constructed from the public datasets (\textit{i.e.} LLVIP~\cite{jia2021llvip}, M3FD~\cite{liu2022target}, MSRS~\cite{tang2022piafusion}, FLIR~\cite{FLIR}) collected in the real scene. It contains a variety of uniformly distributed scenes (daylight, night, rain, foggy, and snow), simulating real environments through the combination of different scenes. Furthermore, we provide the corresponding ground-truth images for each visible image affected by adverse weather conditions for image fusion and image restoration network training. As shown in TABLE~\ref{Dataset}, compared with previous visible-infrared datasets, SWVI is the first one that \add{explicitly investigates how adverse weather conditions affect detection performance}. 

Specifically, we have constructed the dataset from public visible-infrared datasets as follows:
\begin{equation}
    \mathcal{D}_{rain}(J(\bm{\mathrm{X}}))=J(\bm{\mathrm{X}})(1-M_r(\bm{\mathrm{X}}))+R(\bm{X})M_r(\bm{\mathrm{X}}),
\end{equation}
\begin{equation}
    \mathcal{D}_{snow}(J(\bm{\mathrm{X}}))=J(\bm{\mathrm{X}})(1-M_s(\bm{\mathrm{X}}))+S(\bm{\mathrm{X}})M_s(\bm{\mathrm{X}}),
\end{equation}
\begin{equation}
    \mathcal{D}_{foggy}(J(\bm{\mathrm{X}}))=J(\bm{\mathrm{X}})e^{-\int_0^{d(\bm{\mathrm{X}})}\beta dl}+\int_0^{d(\bm{\mathrm{X}})}L_\infty\beta e^{-\beta l} dl,
\end{equation}
where $\bm{\mathrm{X}}$ represents the spatial location in an image,  $\mathcal{D}_{rain}(J(\bm{\mathrm{X}}))$, $\mathcal{D}_{snow}(J(\bm{\mathrm{X}})$ and $\mathcal{D}_{foggy}(J(\bm{\mathrm{X}}))$ represent a function that maps a clear image to one with rain, snow, and fog particle effects, $J(\bm{\mathrm{X}})$ represents the clear image with no weather effects, $M_r(\bm{\mathrm{X}})$ and $M_s(\bm{\mathrm{X}})$ represent rain and snow equivalents, $R(\bm{\mathrm{X}})$ represents a map of the rain masks, $S(\bm{\mathrm{X}})$ represents a chromatic aberration map of the snow particles. Considering scattering effects, $d(\bm{\mathrm{X}})$ represents the distance from the observer at a pixel location $\bm{\mathrm{X}}$, $\beta$ is an atmospheric attenuation coefficient, and $L_\infty$ is the radiance of light. \add{These equations effectively characterize common weather phenomena. For instance, during foggy conditions, the image center is typically more affected by fog, with the effect gradually decreasing towards the periphery. In cases of rain and snow, precipitation often follows a downward trajectory in images~\cite{valanarasu2022transweather}. 
TABLE~\ref{SWVI_construct} presents the methods used for constructing different weather conditions, as well as the corresponding influencing factors considered. 
According to our evaluation, the SWVI dataset achieves Frechet Inception Distance (FID)~\cite{heusel2017gans} and Kernel Inception Distance (KID)~\cite{binkowski2018demystifying} scores of 2.376 and 0.19, respectively, when compared with real-world weather data, demonstrating its strong capability in simulating realistic weather conditions.}

\begin{table*}[!t]
\centering
\setlength{\tabcolsep}{12pt} 
\renewcommand{\arraystretch}{1.1} 
\caption{Quantitative comparisons in terms of PSNR and SSIM (higher is better) with state-of-the-art image deraining, dehazing, and desnowing methods. For the sake of fairness, we uniformly use the visible images from the established SWVI dataset as the evaluation dataset.}
\resizebox{\linewidth}{!}{
\begin{tabular}{lcc lcc lcc}
\toprule
\multirow{2}{*}{\textbf{Image-Deraining}} & \multicolumn{2}{c}{\textbf{SWVI-rain (RGB)}} & \multirow{2}{*}{\textbf{Image-Dehazing}} & \multicolumn{2}{c}{\textbf{SWVI-foggy (RGB)}} & \multirow{2}{*}{\textbf{Image-Desnowing}} & \multicolumn{2}{c}{\textbf{SWVI-snow (RGB)}} \\
\cmidrule(lr){2-3} \cmidrule(lr){5-6} \cmidrule(lr){8-9}
& \textbf{PSNR$\uparrow$} & \textbf{SSIM$\uparrow$} & & \textbf{PSNR$\uparrow$} & \textbf{SSIM$\uparrow$} & & \textbf{PSNR$\uparrow$} & \textbf{SSIM$\uparrow$} \\
\midrule
CycleGAN & 17.65 & 0.7270 & pix2pix & 25.12 & 0.8359 & SPANet & 29.92 & 0.8260 \\
PCNet & 27.13 & 0.6452 & DuRN & 31.44 & 0.9256 & DDMSNet & 34.87 & 0.9462 \\
MPRNet & 29.14 & 0.8546 & AttentiveGAN & 32.56 & 0.9331 & DesnowNet & 32.15 & 0.9416 \\
\add{ESTINet} &\add{34.52} & \add{0.9289} & IDT & 34.14 & 0.9412 & RESCAN & 15.57 & 0.9003 \\
\rowcolor{darkgray!15} \textbf{de-rain (ours)} & \textbf{36.78} & \textbf{0.9464} & \textbf{de-haze (ours)} & \textbf{36.53} & \textbf{0.9795} & \textbf{de-snow (ours)} & \textbf{42.23} & \textbf{0.9821} \\
\midrule
All-in-One & 25.13 & 0.8856 & All-in-One & 31.24 & 0.9122 & All-in-One & 28.12 & 0.8815 \\
TransWeather & 29.77 & 0.9107 & TransWeather & 33.85 & 0.9388 & TransWeather & 35.15 & 0.9394 \\
\add{WeatherDiff} & \add{32.93} & \add{0.9207} & \add{WeatherDiff} & \add{35.36} & \add{0.9598} & \add{WeatherDiff} & \add{37.72} & \add{0.9503} \\
\add{GridFormer} & \add{34.46} & \add{0.9281} & \add{GridFormer} & \add{34.17} & \add{0.9572} & \add{\textbf{GridFormer}} & \add{\textbf{41.10}} & \add{\textbf{0.9617}} \\
\rowcolor{darkgray!15} \textbf{PADM (ours)} & \textbf{35.02} & \textbf{0.9322} & PADM (ours) & 35.88 & 0.9602 & \textbf{PADM (ours)} & \textbf{40.98} & \textbf{0.9578} \\
\bottomrule
\end{tabular}}
\label{quantitative_comparison}
\end{table*}

\begin{table}[!htbp]
    \centering
    \setlength{\tabcolsep}{5pt} 
    \renewcommand{\arraystretch}{1.1} 
    \caption{Comparison of performances with other networks on the LLVIP dataset.}
    \resizebox{\linewidth}{!}{
    \begin{tabular}{lllccc}
        \toprule
        \textbf{Model} & \textbf{Data} & \textbf{Backbone} & \textbf{mAP50$\uparrow$} & \textbf{mAP75$\uparrow$} & \textbf{mAP$\uparrow$} \\
        \midrule
        \multicolumn{6}{c}{\textbf{Mono-modality networks}} \\
        \midrule
        Faster R-CNN & RGB & ResNet50 & 91.4 & 48.0 & 49.2 \\
        Faster R-CNN & Thermal & ResNet50 & 96.1 & 68.5 & 61.1 \\
        DDQ DETR & RGB & ResNet50 & 86.1 & 55.2 & 46.7 \\
        DDQ DETR & Thermal & ResNet50 & 93.9 & 68.8 & 58.6 \\
        SDD & RGB & VGG16 & 82.6 & 31.8 & 39.8 \\
        SDD & Thermal & VGG16 & 90.2 & 57.9 & 53.5 \\
        YOLOv3 & RGB & Darknet53 & 85.9 & 37.9 & 43.3 \\
        YOLOv3 & Thermal & Darknet53 & 89.7 & 53.4 & 52.8 \\
        YOLOv5 & RGB & CSPD53 & 90.8 & 51.9 & 50.0 \\
        YOLOv5 & Thermal & CSPD53 & 94.6 & 70.2 & 61.9 \\
        YOLOv7 & RGB & CSPD53 & 91.4 & 58.4 & 53.6 \\
        YOLOv7 & Thermal & CSPD53 & 94.6 & 70.6 & 62.4 \\
        YOLOv8 & RGB & CSPD53 & 91.9 & 57.7 & 54.0 \\
        YOLOv8 & Thermal & CSPD53 & 95.2 & 72.1 & 62.1 \\
        \add{YOLOv10} & \add{RGB} & \add{CSPD53} & \add{92.2} & \add{58.0} & \add{54.9} \\
        \add{YOLOv10} & \add{Thermal} & \add{CSPD53} & \add{95.3} & \add{72.4} & \add{62.5} \\
        \add{DETR} & \add{RGB} & \add{ResNet50} & \add{89.5} & \add{50.4} & \add{48.1} \\
        \add{DETR} & \add{Thermal} & \add{ResNet50} & \add{93.0} & \add{69.2} & \add{60.5} \\
        \add{Deformable DETR} & \add{RGB} & \add{ResNet50} & \add{91.3} & \add{56.5} & \add{53.8} \\
        \add{Deformable DETR} & \add{Thermal} & \add{ResNet50} & \add{94.5} & \add{70.2} & \add{61.5} \\
        \midrule
        \multicolumn{6}{c}{\textbf{Multi-modality networks}} \\
        \midrule
        GAFF & RGB+T & ResNet18 & 94.0 & 68.8 & 55.8 \\
        ProEN & RGB+T & ResNet50 & 93.4 & 67.3 & 53.5 \\
        CSAA & RGB+T & ResNet50 & 94.3 & 69.5 & 59.2 \\
        RSDet & RGB+T & ResNet50 & 95.8 & 70.9 & 61.3 \\
        DIVFusion & RGB+T & CSPD53 & 89.8 & 63.2 & 52.0 \\
        YOLOv5 & RGB+T & CSPD53 & 95.5 & 70.4 & 62.3 \\
        \add{YOLOv7} & \add{RGB+T} & \add{CSPD53} & \add{95.7} & \add{71.8} & \add{62.6} \\
        \add{YOLOv8} & \add{RGB+T} & \add{CSPD53} & \add{95.6} & \add{71.5} & \add{62.3} \\
        \add{YOLOv10} & \add{RGB+T} & \add{CSPD53} & \add{96.1} & \add{72.7} & \add{63.4} \\
        \add{DETR} & \add{RGB+T} & \add{ResNet50} & \add{93.3} & \add{67.6} & \add{58.5} \\
        \add{Deformable DETR} & \add{RGB+T} & \add{ResNet50} & \add{95.2} & \add{70.1} & \add{60.8} \\
        CFT & RGB+T & CFB & 97.5 & 72.9 & 63.6 \\

        \rowcolor{gray!15} \textbf{CFMW (ours)} & \textbf{RGB+T} & \textbf{CFSSM} & \textbf{98.8}  & \textbf{77.2}  & \textbf{69.8}  \\
        \bottomrule
    \end{tabular}}
    \label{LLVIP}
\end{table}
We divide SWVI into the training set ($34,280$ images), validation set ($17,140$ images), and test set ($8,570$ images). Each folder contains three parts: pairs of visible-infrared images and corresponding weather-influenced visible images. Notice that weather-influenced visible images contain three kinds of weather conditions, classified as SWVI-snow, SWVI-rain, and SWVI-foggy. During the training period, we use the pairs of images (weather-influenced and ground-truth) to train PADM in the first stage, then use the pairs of images (ground-truth and infrared) with corresponding labels to train CFM in the second stage. During the validation and testing period, we use the pairs of images (weather-influenced and infrared) directly, verifying and testing the performance of CFMW under real conditions. We also use this approach when evaluating other methods.

\begin{table}[!t]
    \centering
    \renewcommand{\arraystretch}{1.1} 
    \setlength{\tabcolsep}{5pt} 
    \caption{Comparison of performances with other networks on the SWVI dataset.}
    \resizebox{\linewidth}{!}{
    \begin{tabular}{lllccc}
        \toprule
        \textbf{Model} & \textbf{Data} & \textbf{Backbone} & \textbf{mAP50$\uparrow$} & \textbf{mAP75$\uparrow$} & \textbf{mAP$\uparrow$} \\
        \midrule
        \multicolumn{6}{c}{\textbf{Mono-modality networks}} \\
        \midrule
        Faster R-CNN & RGB & ResNet50 & 82.3 & 34.6 & 30.7 \\
        Faster R-CNN & Thermal & ResNet50 & 90.6 & 63.7 & 55.4 \\
        SDD & RGB & VGG16 & 73.6 & 37.8 & 38.6 \\
        SDD & Thermal & VGG16 & 88.6 & 55.6 & 50.2 \\
        YOLOv3 & RGB & Darknet53 & 78.3 & 29.4 & 24.4 \\
        YOLOv3 & Thermal & Darknet53 & 84.6 & 50.7 & 47.4 \\
        YOLOv5 & RGB & CSPD53 & 80.7 & 38.2 & 30.7 \\
        YOLOv5 & Thermal & CSPD53 & 90.5 & 65.2 & 57.6 \\
        YOLOv7 & RGB & CSPD53 & 85.3 & 41.8 & 34.9 \\
        YOLOv7 & Thermal & CSPD53 & 91.8 & 67.6 & 60.4 \\
        YOLOv8 & RGB & CSPD53 & 86.4 & 42.4 & 36.6 \\
        YOLOv8 & Thermal & CSPD53 & 92.6 & 68.5 & 60.7 \\
        \add{YOLOv10} & \add{RGB} & \add{CSPD53} & \add{88.5} & \add{44.8} & \add{38.1} \\
        \add{YOLOv10} & \add{Thermal} & \add{CSPD53} & \add{94.7} & \add{70.9} & \add{62.7} \\
        \add{DETR} & \add{RGB} & \add{ResNet50} & \add{79.2} & \add{35.4} & \add{31.5} \\
        \add{DETR} & \add{Thermal} & \add{ResNet50} & \add{89.3} & \add{60.5} & \add{54.2} \\
        \add{Deformable DETR} & \add{RGB} & \add{ResNet50} & \add{81.1} & \add{37.3} & \add{33.2} \\
        \add{Deformable DETR} & \add{Thermal} & \add{ResNet50} & \add{92.6} & \add{69.4} & \add{60.7} \\
        \midrule
        \multicolumn{6}{c}{\textbf{Multi-modality networks}} \\
        \midrule
        \add{CSAA} & \add{RGB+T} & \add{ResNet50} & \add{88.3} & \add{63.5} & \add{54.2} \\
        YOLOv5 & RGB+T & CSPD53 & 91.2 & 64.4 & 57.3 \\
        \add{YOLOv7} & \add{RGB+T} & \add{CSPD53} & \add{91.8} & \add{67.4} & \add{58.1} \\
        \add{YOLOv8} & \add{RGB+T} & \add{CSPD53} & \add{91.9} & \add{67.6} & \add{58.7} \\
        \add{YOLOv10} & \add{RGB+T} & \add{CSPD53} & \add{92.1} & \add{68.2} & \add{59.3} \\
        \add{DETR} & \add{RGB+T} & \add{ResNet50} & \add{85.7} & \add{60.1} & \add{56.9} \\
        \add{Deformable DETR} & \add{RGB+T} & \add{ResNet50} & \add{90.2} & \add{67.2} & \add{57.8} \\
        CFT & RGB+T & CFB & 94.4 & 69.7 & 59.4 \\
        \rowcolor{gray!15} \textbf{CFMW (ours)} & \textbf{RGB+T} & \textbf{CFSSM} & \textbf{97.2}  & \textbf{75.9}  & \textbf{68.4}  \\
        \bottomrule
    \end{tabular}}
    \label{SWVI}
\end{table}

\begin{figure*}[!htbp]
    \centering
    \includegraphics[width=\linewidth]{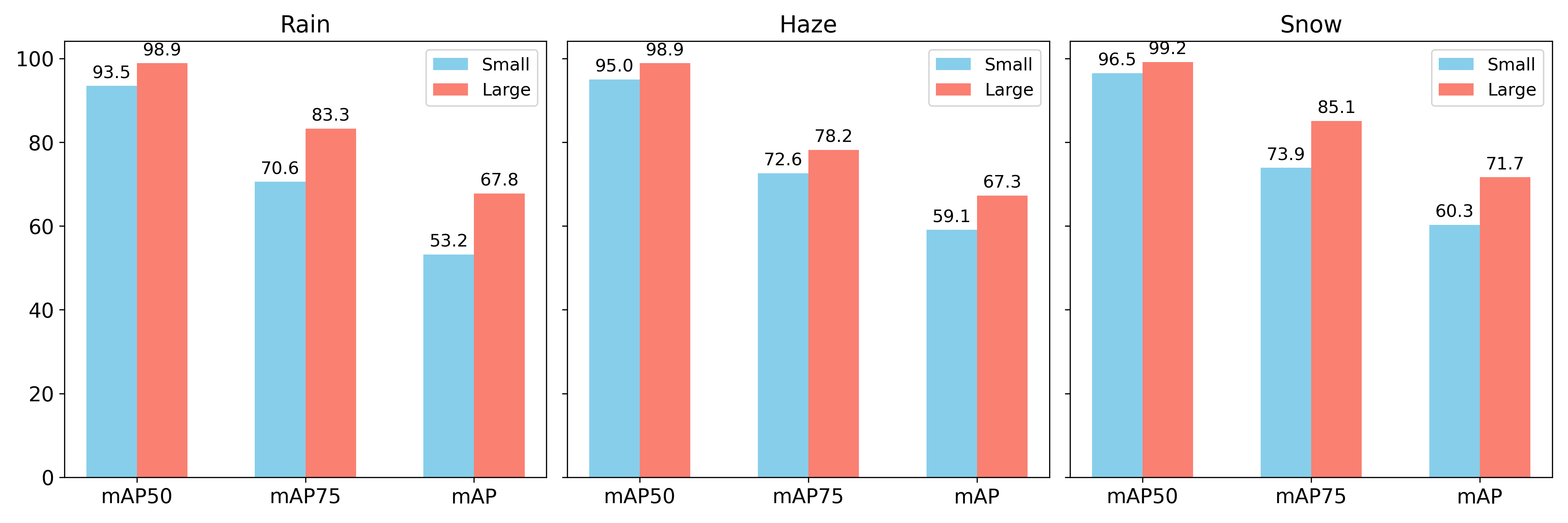}
    \caption{\add{Visualization of accuracy across object sizes and weather conditions. Here, we define the classification criterion for object size based on whether the area of its bounding box is smaller than $2500$. The evaluation metrics were computed separately for large and small objects to provide a more fine-grained analysis of model performance.}}
    \label{visualize_accuracy}
\end{figure*}

\noindent\textbf{Evaluation metrics}. 
We adopt the conventional peak signal-to-noise ratio (PSNR)~\cite{HuynhThu2008ScopeOV} and structural similarity (SSIM)~\cite{1284395} for quantitative evaluations between ground truth and restored images. PSNR is mainly used to evaluate the degree of distortion after image processing, while SSIM pays more attention to the structural information and visual quality of the images. As for object detection quantitative experiments, we introduced three object detection metrics: mean Average Precision (mAP, mAP50, and mAP75) to evaluate the accuracy of the object detection models. 

\textbf{PSNR} could be calculated as follows:
\begin{equation}
    PSNR=10\times lg(\frac{(2^n-1)^2}{MSE}),  
\end{equation}
\begin{equation}
    MSE=\frac{1}{H\times W}\sum_{i=1}^H\sum_{j=1}^W(X(i,j)-Y(i,j))^2,
\end{equation}
where $H$ and $W$ represent the height and width of the images, $n$ is the number of bits per pixel (generally taken as $8$), $X(i,j)$ and $Y(i,j)$ respectively represent the pixel values at the corresponding coordinates.

\noindent \textbf{SSIM} could be calculated as follows:
\begin{equation}
    SSIM=[l(x,y)]^\alpha\cdot[c(x,y)]^\beta\cdot[s(x,y)]^\gamma,
\end{equation}
\begin{equation}
    l(x,y)=\frac{2\mu_x\mu_y+C_1}{\mu_x^2+\mu_y^2+C_1},
\end{equation}
\begin{equation}
    c(x,y)=\frac{2\sigma_x\sigma_y+C_2}{\sigma_x^2+\sigma_y^2+C_2},
\end{equation}
\begin{equation}
    s(x,y)=\frac{\sigma_{xy}+C_3}{\sigma_x\sigma_y+C_3},
\end{equation}
where $l(x,y)$ measures brightness, $c(x,y)$ measures contrast ratio,  $s(x,y)$ measures structure, $\mu$ and $\sigma$ represents mean and standard deviation. $C_1$, $C_1$ and $C_1$ are constants to prevent division by $0$.

\begin{figure}[!t]
    \begin{subfigure}{\linewidth}
        \centering        \includegraphics[width=0.95\linewidth]{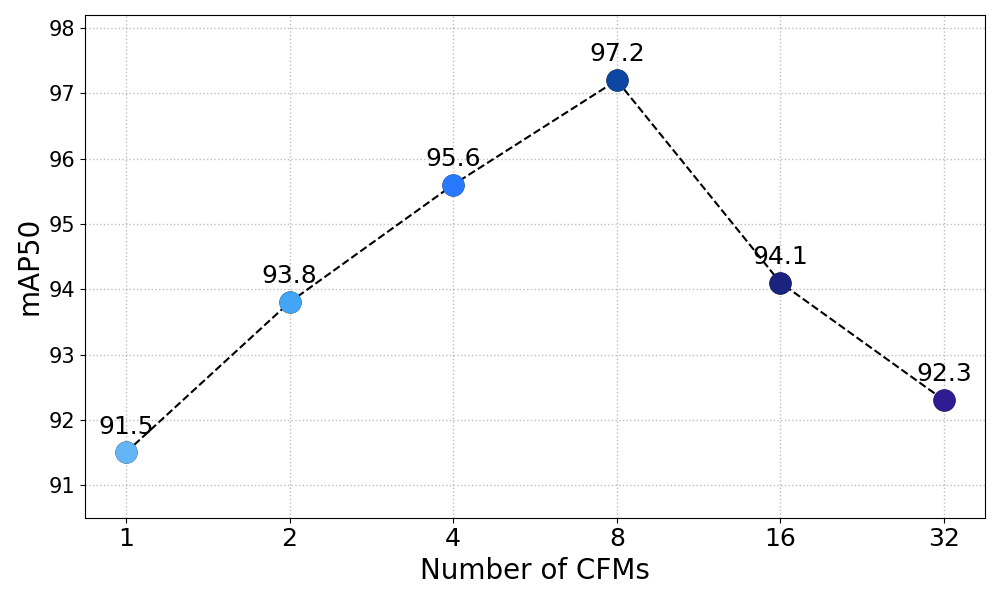}
    \end{subfigure}
    \caption{Ablation study on the number of CFM blocks. The indicators above the circles represent the detection performance of the model under different numbers. \add{The performance of the module on the SWVI dataset was evaluated using mAP50 metrics, where higher values indicate better results.}}
    \label{ablation_num}
\end{figure}

\begin{figure}[!htbp]
    \begin{subfigure}{\linewidth}
        \centering        \includegraphics[width=\linewidth]{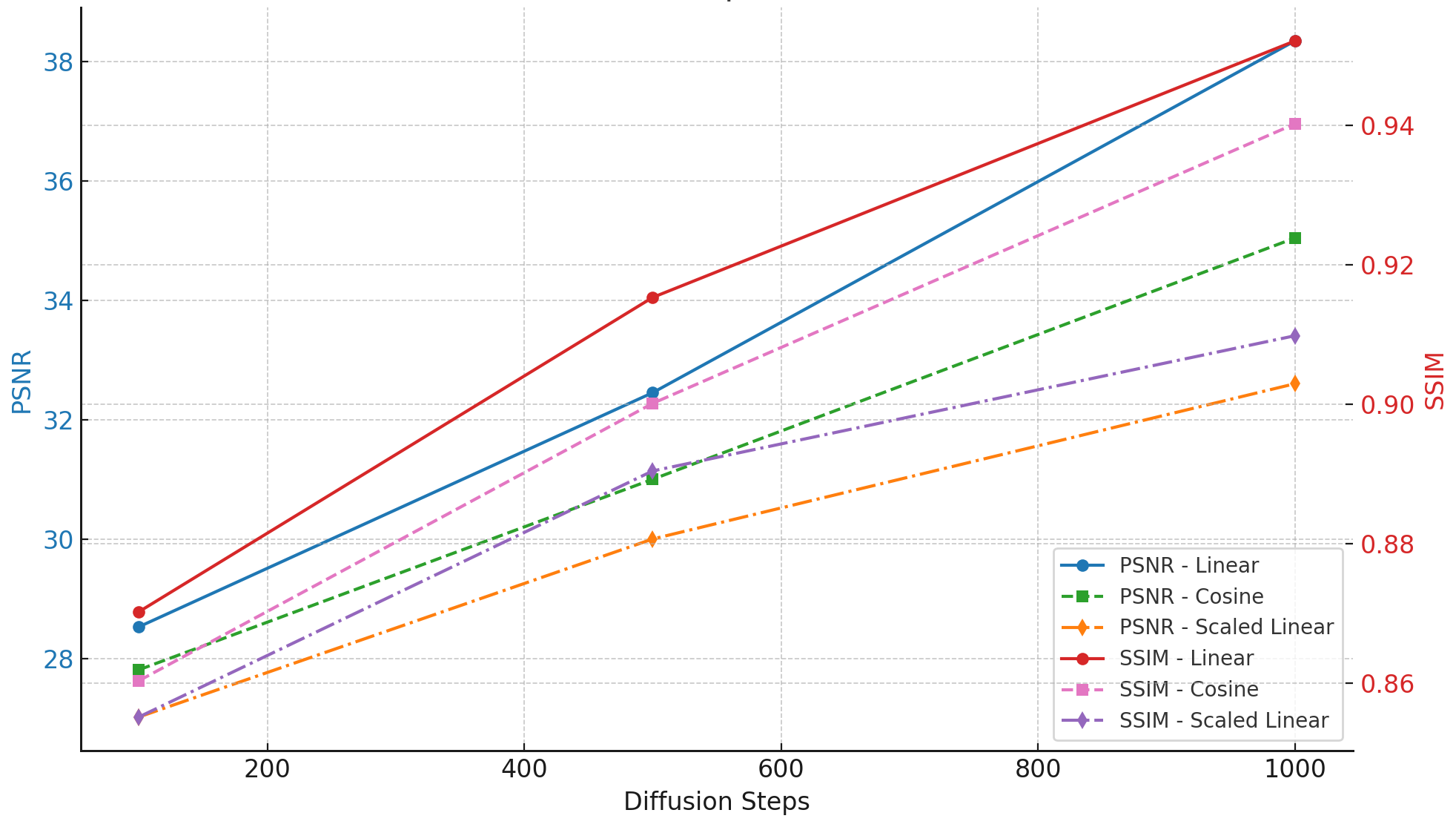}
    \end{subfigure}
    \caption{\add{Ablation study on the hyperparameters of PADM module. We experimented with various noise scheduling strategies (\textit{e.g.}, linear, scaled linear, and cosine) and different diffusion steps (\textit{e.g.}, $100$, $500$, and $1000$). The performance of PADM on the SWVI dataset was evaluated using PSNR~\cite{HuynhThu2008ScopeOV} and SSIM~\cite{1284395} metrics, where higher values indicate better results.}}
    \label{diffusion_ablation}
\end{figure}

\noindent \textbf{mAP, mAP50, and mAP75} could be calculated as follows:
\begin{equation}
    mAP=\frac{1}{n}\sum_{i=1}^NAP_i,
\end{equation}
\begin{equation}
    AP_i=\intop_0^1\mathrm{Precision}\; d(\mathrm{Recall}),
\end{equation}
It should be noted that mAP50 computes the mean of all the AP values for all categories at $\mathrm{IoU}=0.50$, and mAP75 computes the mean at $\mathrm{IoU}=0.75$, similarly.

\begin{figure*}[!htbp]
    \begin{subfigure}{\linewidth}
        \centering        \includegraphics[width=\linewidth]{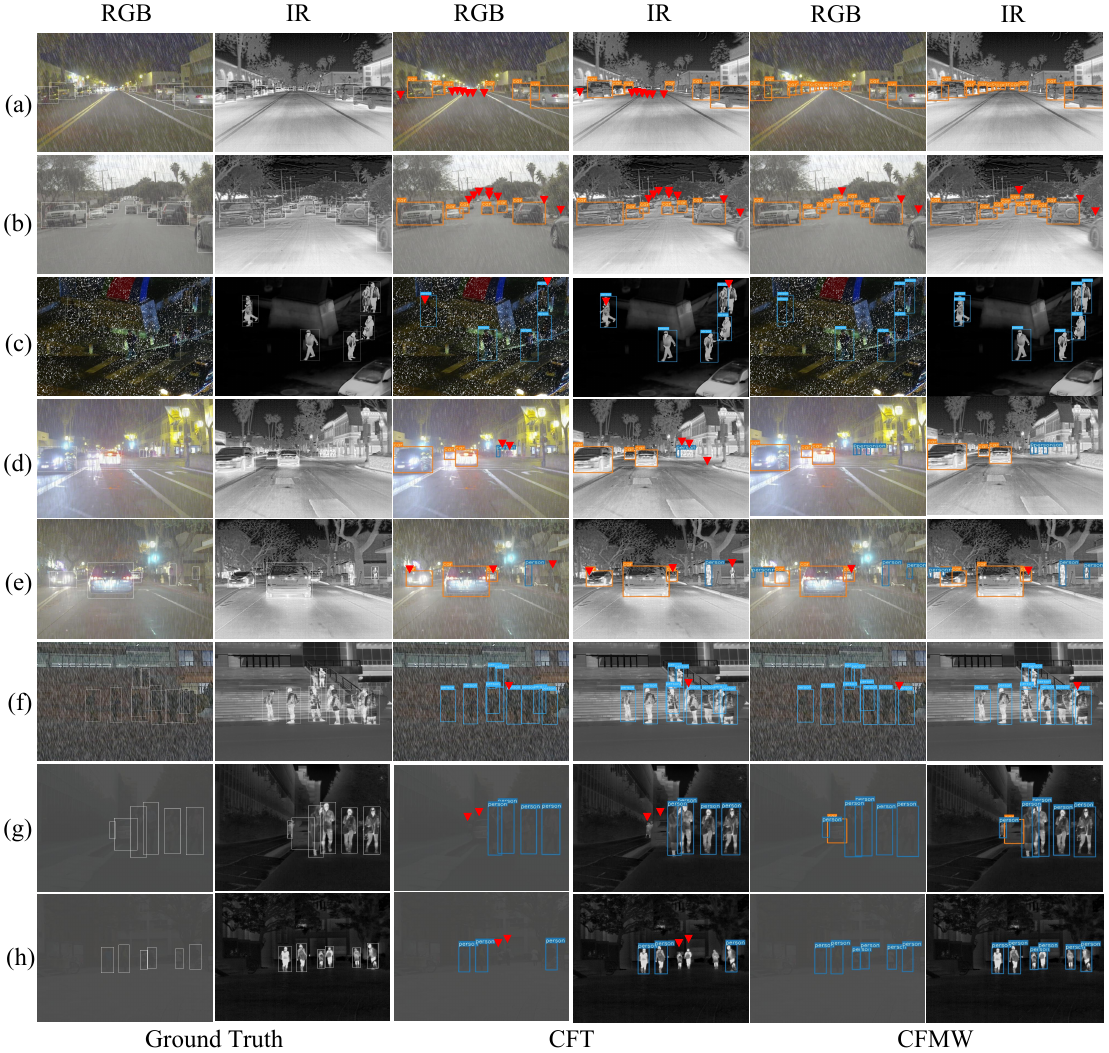}
    \end{subfigure}
    \caption{Examples of daylight and night scenes on the SWVI dataset for visualization. The white box in the figure indicates the ground truth. The inverted triangle indicates the FN samples. Zoom in for more details.}
    \label{comprehensive}
\end{figure*}

\begin{table*}[!htbp]
\centering 
\caption{Ablation experiments on SWVI dataset. To present the general effectiveness of our CFMW, we further combine the PADM and CFM module with other classical detectors (\textit{i.e.}, YOLOv7~\cite{wang2023yolov7}, YOLOv5~\cite{Jocher_YOLOv5_by_Ultralytics_2020}, YOLOv3~\cite{Redmon2018YOLOv3AI}, and Faster R-CNN~\cite{Ren2015FasterRT}). Here \textcolor{black}{\ding{51}} means available while \textcolor{black}{} means unavailable.}
\setlength{\tabcolsep}{8pt} 
\renewcommand{\arraystretch}{1.3} 
\resizebox{\linewidth}{!}{
\begin{tabular}{@{}lllllllllll@{}}
\hline
Modality   & Method             & Detector       & \multicolumn{1}{c}{Shallow Swapping} & \multicolumn{1}{c}{CFSSM} & \multicolumn{1}{c}{PADM} & mAP50$\uparrow$ & mAP75$\uparrow$ & mAP$\uparrow$ \\ 
\hline
RGB        & CSPDarknet53       & YOLOv7         & \multicolumn{1}{c}{\textcolor{black}{}} & \multicolumn{1}{c}{\textcolor{black}{}} & \multicolumn{1}{c}{\textcolor{black}{}} & 85.3  & 41.8  & 34.9  \\
Thermal    & CSPDarknet53       & YOLOv7         & \multicolumn{1}{c}{\textcolor{black}{}} & \multicolumn{1}{c}{\textcolor{black}{}} & \multicolumn{1}{c}{\textcolor{black}{}} & 92.8  & 72.6  & 60.4  \\
RGB+T      & +Two stream        & YOLOv7         & \multicolumn{1}{c}{\textcolor{black}{}} & \multicolumn{1}{c}{\textcolor{black}{}} & \multicolumn{1}{c}{\textcolor{black}{}} & 92.4  & 65.1  & 60.4  \\
RGB+T      & +CFM              & YOLOv7         & \multicolumn{1}{c}{\textcolor{black}{\ding{51}}} & \multicolumn{1}{c}{\textcolor{black}{}} & \multicolumn{1}{c}{\textcolor{black}{}} & 93.8  & 65.8  & 62.8  \\
RGB+T      & +CFM              & YOLOv7         & \multicolumn{1}{c}{\textcolor{black}{}} & \multicolumn{1}{c}{\textcolor{black}{\ding{51}}} & \multicolumn{1}{c}{\textcolor{black}{}} & 94.2  & 66.2  & 63.1  \\
RGB+T      & +CFM              & YOLOv7         & \multicolumn{1}{c}{\textcolor{black}{\ding{51}}} & \multicolumn{1}{c}{\textcolor{black}{\ding{51}}} & \multicolumn{1}{c}{\textcolor{black}{}} & 95.4  & 68.2  & 63.9  \\
RGB+T      & +PADM             & YOLOv7         & \multicolumn{1}{c}{\textcolor{black}{}} & \multicolumn{1}{c}{\textcolor{black}{}} & \multicolumn{1}{c}{\textcolor{black}{\ding{51}}} & 94.5  & 67.9  & 63.8  \\
\rowcolor{gray!15}
RGB+T      & \textbf{+CFSSM\&PADM} & YOLOv7         & \multicolumn{1}{c}{\textcolor{black}{\ding{51}}} & \multicolumn{1}{c}{\textcolor{black}{\ding{51}}} & \multicolumn{1}{c}{\textcolor{black}{\ding{51}}} & \textbf{96.6} & \textbf{75.1} & \textbf{64.1} \\ 
\hline
RGB        & CSPDarknet53       & YOLOv5         & \multicolumn{1}{c}{\textcolor{black}{}} & \multicolumn{1}{c}{\textcolor{black}{}} & \multicolumn{1}{c}{\textcolor{black}{}} & 80.7  & 38.2  & 30.7  \\
Thermal    & CSPDarknet53       & YOLOv5         & \multicolumn{1}{c}{\textcolor{black}{}} & \multicolumn{1}{c}{\textcolor{black}{}} & \multicolumn{1}{c}{\textcolor{black}{}} & 90.5  & 65.2  & 57.6  \\
RGB+T      & +Two stream        & YOLOv5         & \multicolumn{1}{c}{\textcolor{black}{}} & \multicolumn{1}{c}{\textcolor{black}{}} & \multicolumn{1}{c}{\textcolor{black}{}} & 91.2  & 67.4  & 59.3  \\
RGB+T      & +CFM              & YOLOv5         & \multicolumn{1}{c}{\textcolor{black}{\ding{51}}} & \multicolumn{1}{c}{\textcolor{black}{}} & \multicolumn{1}{c}{\textcolor{black}{}} & 91.5  & 63.6  & 60.3  \\
RGB+T      & +CFM              & YOLOv5         & \multicolumn{1}{c}{\textcolor{black}{}} & \multicolumn{1}{c}{\textcolor{black}{\ding{51}}} & \multicolumn{1}{c}{\textcolor{black}{}} & 93.9 & 66.5  & 62.3  \\
RGB+T      & +CFM              & YOLOv5         & \multicolumn{1}{c}{\textcolor{black}{\ding{51}}} & \multicolumn{1}{c}{\textcolor{black}{\ding{51}}} & \multicolumn{1}{c}{\textcolor{black}{}} & 94.8  & 67.6  & 62.9  \\
RGB+T      & +PADM             & YOLOv5         & \multicolumn{1}{c}{\textcolor{black}{}} & \multicolumn{1}{c}{\textcolor{black}{}} & \multicolumn{1}{c}{\textcolor{black}{\ding{51}}} & 95.4  & 68.2  & 62.8  \\
\rowcolor{gray!15}
RGB+T      & \textbf{+CFSSM\&PADM} & YOLOv5         & \multicolumn{1}{c}{\textcolor{black}{\ding{51}}} & \multicolumn{1}{c}{\textcolor{black}{\ding{51}}} & \multicolumn{1}{c}{\textcolor{black}{\ding{51}}} & \textbf{97.2} & \textbf{76.9} & \textbf{63.4} \\ 
\hline
RGB        & Darknet53          & YOLOv3         & \multicolumn{1}{c}{\textcolor{black}{}} & \multicolumn{1}{c}{\textcolor{black}{}} & \multicolumn{1}{c}{\textcolor{black}{}} & 78.3  & 29.4  & 24.4  \\
Thermal    & Darknet53          & YOLOv3         & \multicolumn{1}{c}{\textcolor{black}{}} & \multicolumn{1}{c}{\textcolor{black}{}} & \multicolumn{1}{c}{\textcolor{black}{}} & 84.6  & 50.7  & 47.4  \\
RGB+T      & +Two stream        & YOLOv3         & \multicolumn{1}{c}{\textcolor{black}{}} & \multicolumn{1}{c}{\textcolor{black}{}} & \multicolumn{1}{c}{\textcolor{black}{}} & 91.8  & 63.9  & 55.3  \\
RGB+T      & +CFM              & YOLOv3         & \multicolumn{1}{c}{\textcolor{black}{\ding{51}}} & \multicolumn{1}{c}{\textcolor{black}{}} & \multicolumn{1}{c}{\textcolor{black}{}} & 92.5  & 64.4  & 57.9  \\
RGB+T      & +CFM              & YOLOv3         & \multicolumn{1}{c}{\textcolor{black}{}} & \multicolumn{1}{c}{\textcolor{black}{\ding{51}}} & \multicolumn{1}{c}{\textcolor{black}{}} & 94.9  & 66.2  & 59.9  \\
RGB+T      & +CFM              & YOLOv3         & \multicolumn{1}{c}{\textcolor{black}{\ding{51}}} & \multicolumn{1}{c}{\textcolor{black}{\ding{51}}} & \multicolumn{1}{c}{\textcolor{black}{}} & 96.1  & 68.6  & 61.4  \\
RGB+T      & +PADM             & YOLOv3         & \multicolumn{1}{c}{\textcolor{black}{}} & \multicolumn{1}{c}{\textcolor{black}{}} & \multicolumn{1}{c}{\textcolor{black}{\ding{51}}} & 93.5  & 67.3  & 58.2  \\
\rowcolor{gray!15}
RGB+T      & \textbf{+CFSSM\&PADM} & YOLOv3         & \multicolumn{1}{c}{\textcolor{black}{\ding{51}}} & \multicolumn{1}{c}{\textcolor{black}{\ding{51}}} & \multicolumn{1}{c}{\textcolor{black}{\ding{51}}} & \textbf{96.7} & \textbf{70.2} & \textbf{62.6} \\ 
\hline
RGB        & ResNet50           & Faster R-CNN   & \multicolumn{1}{c}{\textcolor{black}{}} & \multicolumn{1}{c}{\textcolor{black}{}} & \multicolumn{1}{c}{\textcolor{black}{}} & 82.3  & 34.6  & 29.8  \\
Thermal    & ResNet50           & Faster R-CNN   & \multicolumn{1}{c}{\textcolor{black}{}} & \multicolumn{1}{c}{\textcolor{black}{}} & \multicolumn{1}{c}{\textcolor{black}{}} & 90.6  & 63.7  & 55.4  \\
RGB+T      & +Two stream        & Faster R-CNN   & \multicolumn{1}{c}{\textcolor{black}{}} & \multicolumn{1}{c}{\textcolor{black}{}} & \multicolumn{1}{c}{\textcolor{black}{}} & 93.5  & 62.8  & 57.1  \\
RGB+T      & +CFM              & Faster R-CNN   & \multicolumn{1}{c}{\textcolor{black}{\ding{51}}} & \multicolumn{1}{c}{\textcolor{black}{}} & \multicolumn{1}{c}{\textcolor{black}{}} & 93.7  & 64.2  & 58.8  \\
RGB+T      & +CFM              & Faster R-CNN         & \multicolumn{1}{c}{\textcolor{black}{}} & \multicolumn{1}{c}{\textcolor{black}{\ding{51}}} & \multicolumn{1}{c}{\textcolor{black}{}} & 95.9  & 68.4  & 60.8  \\
RGB+T      & +CFM              & Faster R-CNN         & \multicolumn{1}{c}{\textcolor{black}{\ding{51}}} & \multicolumn{1}{c}{\textcolor{black}{\ding{51}}} & \multicolumn{1}{c}{\textcolor{black}{}} & 96.2  & 69.1  & 61.3  \\
RGB+T      & +PADM             & Faster R-CNN   & \multicolumn{1}{c}{\textcolor{black}{}} & \multicolumn{1}{c}{\textcolor{black}{}} & \multicolumn{1}{c}{\textcolor{black}{\ding{51}}} & 94.5  & 67.1  & 58.6  \\
\rowcolor{gray!15}
RGB+T      & \textbf{+CFSSM\&PADM} & Faster R-CNN   & \multicolumn{1}{c}{\textcolor{black}{\ding{51}}} & \multicolumn{1}{c}{\textcolor{black}{\ding{51}}} & \multicolumn{1}{c}{\textcolor{black}{\ding{51}}} & \textbf{96.2} & \textbf{69.7} & \textbf{62.2} \\ 
\hline
\end{tabular}
}
\label{SWVI_snow}
\end{table*}

\subsection{Implementation Details}
\noindent As for PADM, we performed experiments both in specific-weather conditions and multi-weather conditions image restoration settings. We denote our specific-weather restoration models as de-rain, de-snow, and de-foggy to verify the general PADM model under specific weather conditions. We trained the $128{\times}128$ patch size version of all models. We use Adam as an optimizer while training all the models we compare. \add{We use the linear noise scheduling strategy. In accordance with the hyperparameters commonly employed in diffusion network designs~\cite{ho2020denoising, song2020denoising, ozdenizci2023}, we set the initial value of $\beta$ to $0.001$ and the final value to $0.02$.
} 
During the training process, we trained PADM for $3{\times}10^6$ iterations with $1000$ diffusion steps for $3$ days on a single RTX A6000 graphics card (48GB RAM). 
As for CFM, we did not perform task-specific parameter tuning or modifications to the network architecture. For better performance, we select the YOLOv5 model's public weight initialization (yolov5l.pt), which is pre-trained on the large-scale COCO dataset~\cite{Lin2014MicrosoftCC}. During the training stage, we set the batch size to $32$, the Adam optimizer is set with a momentum of $0.98$, and the learning rate starts from $0.001$. The loss weight parameters $\lambda_{box}$, $\lambda_{cls}$ and $\lambda_{conf}$ in loss $\mathcal{L}_{total}$ are set to 1.0 and 1.0, respectively.

\begin{figure*}[!htbp]
    \centering
    \includegraphics[width=\linewidth]{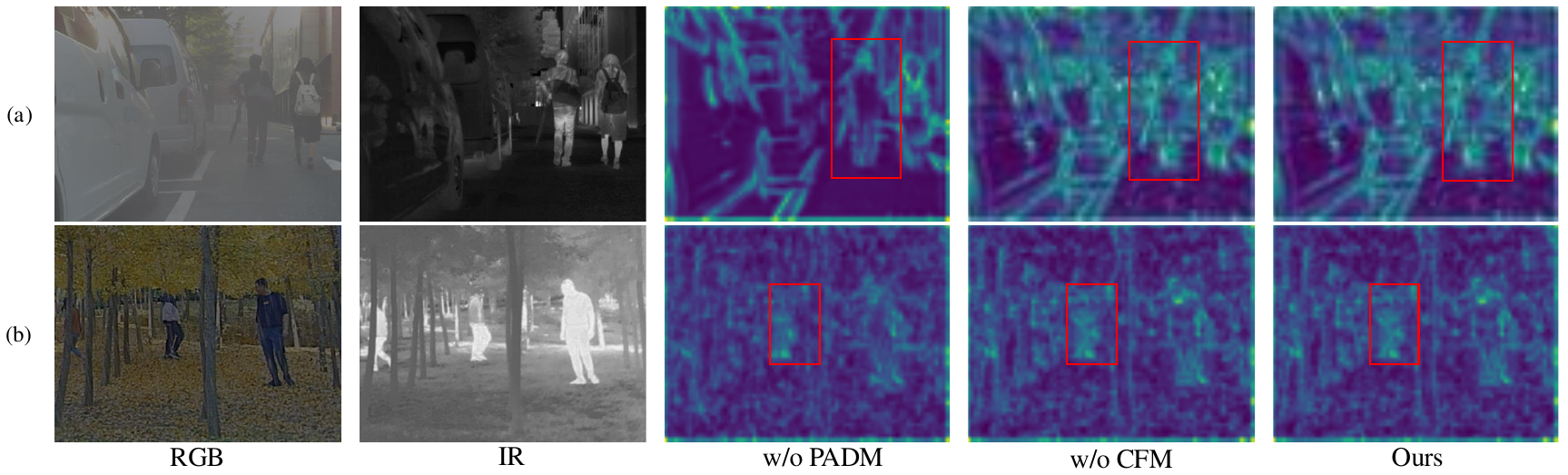}
    \caption{\add{Ablation experiments designed for the PADM and CFM modules are presented, where (a) shows results collected from the SWVI dataset and (b) from the M3FD dataset. The red boxes highlight the features that have a significant impact on the detection performance. Zoom in for more details.}}
    \label{fig6}
\end{figure*}

\begin{figure}[!htbp]
    \begin{subfigure}{\linewidth}
        \centering        \includegraphics[width=\linewidth]{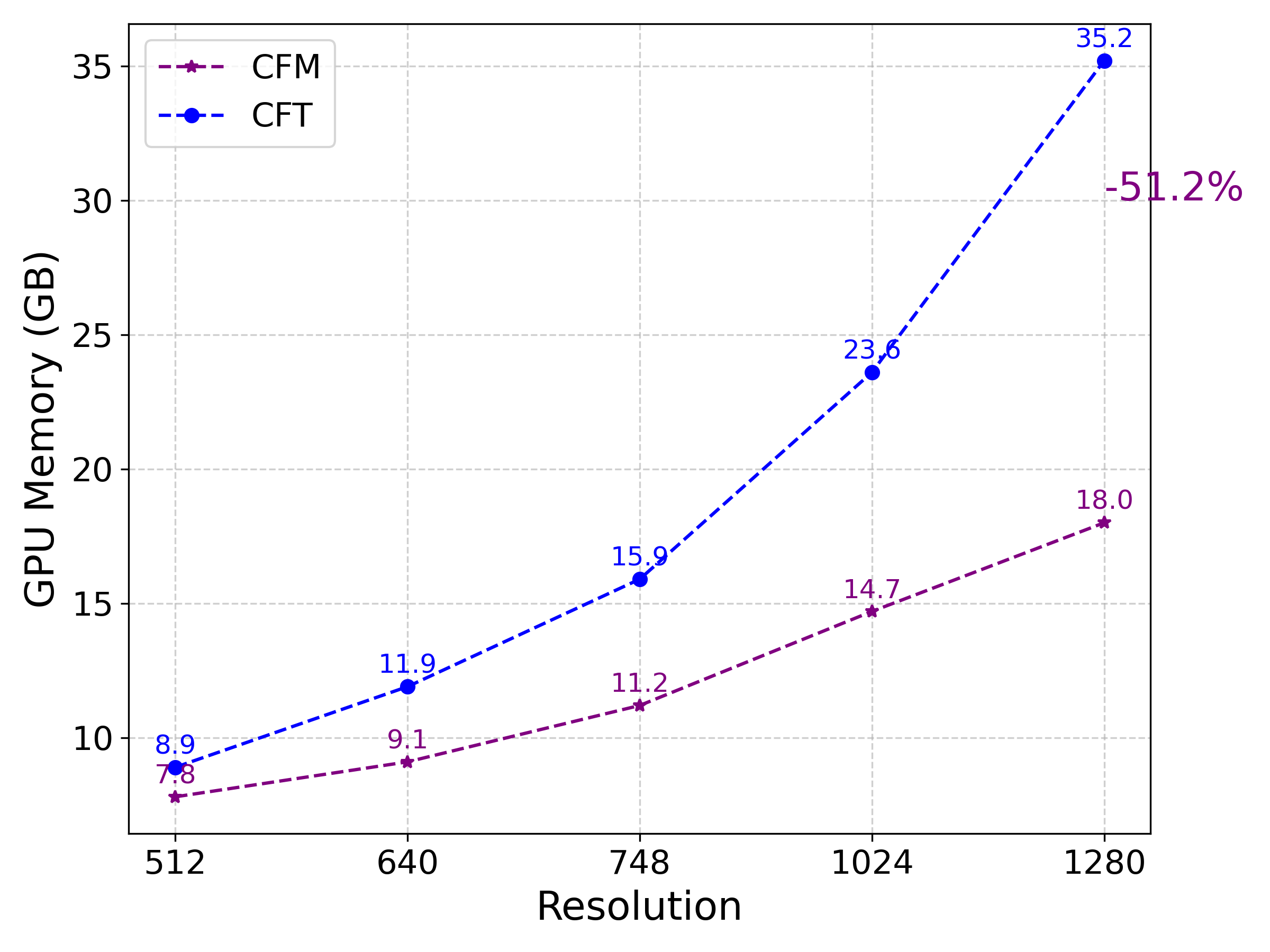}
    \end{subfigure}
    \caption{Ablation study on the efficiency comparison between CFT and CFM. The indicators above the circles represent the GPU usage of the model under different resolutions.}
    \label{ablation_gpu}
\end{figure}

\begin{table*}[!htbp]
    \centering
    \renewcommand{\arraystretch}{1.1} 
    \setlength{\tabcolsep}{10pt}      

    \caption{\add{Comparative results on the SWVI dataset, including input resolution, detection performance, GFLOPs, and FPS.}}
    \resizebox{\linewidth}{!}{
    \begin{tabular}{lcccccc}
        \toprule
        \textbf{Methods} & \textbf{Image Size} & $\mathbf{mAP50}\uparrow$ & $\mathbf{mAP75}\uparrow $ & $\mathbf{mAP}\uparrow$ & \textbf{GFLOPs} & \textbf{FPS} \\
        \midrule
        CFT & $640^2$ & 93.4 & 71.8 & 59.7 & 290.73 & 5.14 \\
        Ours & $640^2$ & 97.2 & 76.9 & 63.4 & 308.62 & 13.52 \\
        \quad w/o Shallow Swapping & $640^2$ & 96.4 & 75.8 & 62.5 & 308.62 & 15.73 \\
        \quad w/o CFSSM & $640^2$ & 94.8 & 72.3 & 60.2 & 95.58 & 19.54 \\
        \quad w/o Shallow Swapping\& CFSSM & $640^2$ & 92.2 & 70.6 & 58.4 & 95.58 & 23.64 \\
        \quad w/ Shared Decoder & $640^2$ & 95.7 & 73.2 & 60.3 & 256.57 & 13.81 \\
        \bottomrule
    \end{tabular}}
    \label{FPS_compare}
\end{table*}

\subsection{Comparative Experiments}
\noindent In this section, we make comparisons with several state-of-the-art methods in image deweathering and cross-modality object detection separately. In TABLE~\ref{quantitative_comparison}, we perform comparisons with methods for image desnowing (\textit{i.e.} SPANet~\cite{Wang2019SpatialAS}, DDMSNet~\cite{Zhang2021DeepDM}, {DesnowNet~\cite{JSTASRChen}}, RESCAN~\cite{li2018recurrent}), deraining (\textit{i.e.} \add{ESTINet~\cite{zhang2022enhanced}}, Cycle-GAN~\cite{Zhu2017UnpairedIT}, PCNet~\cite{Jiang2021RainFreeAR}, MPRNet~\cite{Zamir2021MultiStagePI}), and dehazing (\textit{i.e.} pix2pix~\cite{Isola2016ImagetoImageTW}, DuRN~\cite{Liu2019DualRN}, Attentive-GAN~\cite{Qian2017AttentiveGA}, IDT~\cite{Xiao2022ImageDT}), \add{as well as four state-of-the-art multi-weather image restoration methods: All in One~\cite{li2020all}, TransWeather~\cite{valanarasu2022transweather}, GridFormer~\cite{wang2024gridformer}, and WeatherDiff~\cite{ozdenizci2023}}. In TABLE~\ref{LLVIP} and TABLE~\ref{SWVI},  to prove the consistent improvements of CFMW, we compare with several base single-modality object detection methods (\textit{i.e.}, Faster R-CNN~\cite{Ren2015FasterRT}, SDD~\cite{Liu2015SSDSS}, YOLOv3~\cite{Redmon2018YOLOv3AI}, YOLOv5~\cite{Jocher_YOLOv5_by_Ultralytics_2020}, YOLOv7~\cite{wang2023yolov7}, \add{YOLOv8~\cite{yolov8}, YOLOv10~\cite{wang2024yolov10}, DETR~\cite{carion2020end}, Deformable DETR~\cite{zhu2020deformable}}), and several multi-modality object detection methods (\textit{i.e.}, our baseline, standard two-stream YOLOv5 object detection network, and CFT~\cite{qingyun2021cross}).

\noindent\textbf{Comparison of image deweathering}. 
As shown in TABLE~\ref{quantitative_comparison}, we use the RGB modality of the SWVI dataset (including rain, foggy, and haze weather conditions) as a comparative dataset to measure the performance of different models under different weather conditions. 
The top of the table contains results from specific-weather image restoration, where we set the sampling time steps $S{=}50$. 
For image-deraining, image-dehazing, and image-desnowing tasks, the proposed solution consistently achieves the best results ($36.78/0.9464$ on SWVI-rain, $36.53/0.9795$ on SWVI-foggy, and $42.23/0.9821$ on SWVI-snow). Especially, in the image de-rain task, the performance improvement is about $24\%$ compared with the current state-of-the-art method (MPRNet~\cite{Zamir2021MultiStagePI}). 
For multi-weather image restoration, although the results are not as good as the specific-weather model due to the complexity of the task, the proposed method also reaches the best results ( $35.02/0.9322$ on SWVI-rain, $35.88/0.9602$ on SWVI-foggy, and $40.98/0.9578$ on SWVI-snow) compared with All in One~\cite{li2020all} and TransWeather~\cite{valanarasu2022transweather}, with about $17\%$ performance improvement compared against TransWeather~\cite{valanarasu2022transweather} and $25\%$ performance improvement compared against All in One~\cite{li2020all}.

\noindent\textbf{Comparison of cross-modality object detection}. 
As shown in TABLE~\ref{LLVIP} and TABLE~\ref{SWVI}, we use LLVIP~\cite{jia2021llvip} and SWVI as the comparative datasets. The top of the table contains results from single-modality networks, each of which uses the RGB modality or the thermal modality for detection. The bottom of the table shows results from current SOTA multi-modality networks, including basic two-stream YOLOv5~\cite{Jocher_YOLOv5_by_Ultralytics_2020}, \add{YOLOv7~\cite{wang2023yolov7},  YOLOv8~\cite{yolov8}, YOLOv10~\cite{wang2024yolov10},} CFT~\cite{qingyun2021cross}, ProEN~\cite{chen2022multimodal}, GAFF~\cite{ZHANG2021GuidedAF}, CSAA~\cite{cao2023multimodal}, RSDet~\cite{zhao2024removal}, DIVFusion~\cite{tang2023divfusion}, and the proposed CFMW. According to TABLE~\ref{SWVI}, it can be observed that with the integration of PADM and CFM, CFMW achieves an overwhelming performance improvement on each metric (mAP50: $2.3{\uparrow}$, mAP75: $4.3{\uparrow}$, mAP: $3.0{\uparrow}$) on SWVI-snow compared with the best existing network on each metric, which shows that it has preferable adaptability under adverse weather conditions. Also, CFMW can achieve a more accurate detection (mAP50: $98.8$, mAP75: $77.2$, mAP: $64.8$) with lower computational consumption, as shown in TABLE~\ref{LLVIP}. 

Meanwhile, Fig.~\ref{comprehensive} visualizes the performance of CFMW on the SWVI dataset compared with CFT~\cite{qingyun2021cross}. As can be seen from the figure, compared with CFT based on attention mechanism fusion, CFMW is more robust against weather interference and still maintains a stable detection effect in extreme scenarios such as overlapping multiple targets and small objects. The detection results of CFMW are very close to the ground truth. It is speculated that this is because the addition of PADM reduces the image noise caused by bad weather, while CFM superimposes the fused features to supplement the features of the objects in the picture. However, it is undeniable that in some specific scenarios, CFMW still lacks detection of small objects in the picture, which requires subsequent work to improve this type of special problem and further improve the robustness of the model for multimodal object detection under adverse weather conditions.

\subsection{Ablation Study and Analysis}
\label{ablation study}
\noindent In this section, we analyze the effectiveness of CFMW. We first validate the importance of PADM and CFM modules in performance improvement in a parametric form through detailed ablation experiments. Then, we verify the actual effect of the model by visualizing the features. Finally, we conduct ablation experiments on some hyperparameter settings.

\noindent \textbf{Exploration experiments.} 
\add{As shown in Fig.~\ref{visualize_accuracy}, to analyze how different adverse weather types impact object sizes and detection accuracy, we categorize and analyze the target objects within the SWVI dataset. Here, we define the classification criterion for object size based on whether the area of its bounding box is smaller than $2500$. This threshold is chosen because an object of size $50{\times}50$ is considered relatively difficult to recognize at a resolution of $1280{\times}1024$. Based on this criterion, we classify $86,932$ objects as large, accounting for $73.8\%$, and $30,856$ objects as small, making up $26.2\%$. The findings indicate that the network is highly sensitive to small object detection under all three weather conditions, especially in foggy environments. We speculate that this is due to noise introduced by adverse weather conditions, which causes blurry image edges and reduces the clarity of extracted features, leading to missed detections.}

\noindent\textbf{Hyperparameters experiments.} 
\add{We conducted extensive ablation studies on the hyperparameters of both the PADM and CFSSM modules to investigate under which configurations the CFMW framework achieves optimal performance. As shown in Fig.~\ref{ablation_num}, it can be observed that the overall performance is optimal when the number of CFM blocks reaches $8$. As the number of stacked blocks increases, the model's performance will improve, but too many CFM blocks will cause the model to overfit on a limited data set, which reduces its generalization ability. 
We conducted ablation experiments on PADM to vary the number of diffusion steps and noise injection strategies. The proportion of the original image and the added noise at each time step under these different strategies is illustrated in Fig.~\ref{diffusion_ablation}.  As shown in Fig.~\ref{fig6}, we tested and visualized the impact of removing the PADM module and the CFM modules on cross-modality feature extraction and fusion. It can be observed that after removing the PADM module, the features extracted by the model are greatly reduced, presumably due to the fuzzy noise caused by the weather. Removing the CFM will lead to a decrease in the model's recognition of the contour.}

\noindent\textbf{Qualitative experiments.} To verify the effectiveness of PADM and CFM modules, we visually show the ablation of PADM and CFM. To understand the impact of each component in CFMW, we performed a set of comprehensive ablation experiments. As shown in TABLE~\ref{SWVI_snow}, we further combine the CFM and PADM with other classical detectors, \textit{i.e.}, YOLOv7~\cite{wang2023yolov7}, YOLOv5~\cite{Jocher_YOLOv5_by_Ultralytics_2020}, and Faster R-CNN~\cite{Ren2015FasterRT} to present the general effectiveness of our CFMW. The proposed CFMW improves the performance of cross-modality object detection using either a one-stage or two-stage detector under complex weather conditions. Specifically, CFM achieves an $11.3\%$ gain on mAP50, an $81.6\%$ gain on mAP75, and a $78.3\%$ gain on mAP (on YOLOv5~\cite{Jocher_YOLOv5_by_Ultralytics_2020}). 
After adding PADM, we achieved a $12.1\%$ gain on mAP50, an $88.2\%$ gain on mAP75, and an $80.4\%$ gain on mAP.  

We also conduct a comparison with CFT about efficiency. As shown in Fig.~\ref{ablation_gpu}, we implement this experiment on the SWVI dataset with a batch size of $8$. As the resolution continues to increase, the rate of increase in GPU memory usage during CFT training is much higher than that of CFM. When the resolution reaches $1280\times 1280$, GPU memory required by CFT reaches $35.2GB$, while CFT only needs $18GB$ under the same conditions, saving $51.2\%$. When the resolution exceeds $1280{\times}1280$, the memory capacity required by CFT exceeds the RTX A6000 graphics card RAM ($48GB$ RAM), making it impossible to continue training. \add{In TABLE~\ref{FPS_compare}, we quantitatively investigate the differences between CFM and CFT in terms of FPS. From the results, we observe that our proposed method significantly improves inference speed compared to the baseline CFT~\cite{qingyun2021cross} (13.52 vs. 5.14 FPS), while maintaining superior detection performance. As shown, under the same settings, CFM is nearly $3$ times faster than CFT~\cite{qingyun2021cross}, demonstrating its efficiency and effectiveness in cross-modality object detection.}

\section{conclusion and future work}
\noindent \textbf{Conclusion.} 
In this work, we introduce a novel approach to visible-infrared object detection under severe weather conditions, namely the Severe Weather Visible-Infrared Dataset (SWVI). 
We provide a valuable resource for training and evaluating models in realistic and challenging environments. The Cross-modality Fusion Mamba with Weather-robust (CFMW) proves to be highly effective in enhancing detection accuracy while managing computational efficiency. Extensive experiments show that CFMW outperforms existing benchmarks, achieving state-of-the-art. 
This work opens up new possibilities for cross-modality object detection under adverse weather.

\noindent\textbf{Future work.} 
Visible-infrared data is usually captured by dedicated equipment. Common application scenarios of such equipment include video surveillance, security, and autonomous driving, which have high requirements for the quality of collected data and the accuracy of intelligent recognition.  Meanwhile, adverse weather is also very common in such scenarios. Unfortunately, current attention paid to such issues is still not very high, and there is a lack of corresponding research paradigms and data resources. In this work, we proposed a conventional solution and provided corresponding data to verify the effectiveness of our proposed solution. In the future, we hope that more work can focus on this issue, collect more real visible-infrared data affected by weather, and propose solutions with higher recognition accuracy, higher computational efficiency, and simpler model architecture.

\noindent\textbf{Acknowledgements.} This work was conducted on the Earth System Big Data Platform of the School of Earth Sciences, Zhejiang University.

\bibliographystyle{IEEEtran}
\bibliography{IEEEabrv,reference}

\end{document}